\documentclass[a4paper,fleqn]{cas-dc}

\usepackage[numbers]{natbib}

\usepackage{xspace}
\usepackage{graphicx}
\usepackage{amssymb}
\usepackage{amsmath}
\usepackage{extarrows}
\usepackage{dsfont}
\usepackage{multirow}
\usepackage{float}
\usepackage{nicefrac}
\usepackage{circledsteps}
\usepackage{colortbl}
\usepackage[dvipsnames, table]{xcolor}
\usepackage{booktabs}
\usepackage{siunitx}
\usepackage[capitalize]{cleveref}
\usepackage[figuresleft]{rotating}
\usepackage{tablefootnote}
\usepackage{pifont}
\usepackage{makecell}
\usepackage{tikz}
\usepackage{algorithm}
\usepackage[noend]{algpseudocode}
\usepackage{tcolorbox}
\usepackage{soul}
\usepackage{subcaption}
\usepackage{enumitem}

\definecolor{lightgray}{gray}{0.95}
\definecolor{midgray}{gray}{0.55}
\definecolor{steelblue}{HTML}{4D82B7}
\definecolor{davysgrey}{rgb}{0.33, 0.33, 0.33}
\definecolor{LightCyan}{rgb}{0.88,1,1}
\definecolor{ao(english)}{rgb}{0.0, 0.5, 0.0}
\definecolor{lightblue}{rgb}{0.9, 0.95, 1.0}
\definecolor{warmup}{HTML}{f7d779}
\definecolor{specialization}{HTML}{9fc5fc}
\definecolor{softyellow}{RGB}{255, 255, 204}

\newcommand{\Star}[1]{#1\ensuremath{^*}\kern-\scriptspace}

\newcommand{\tit}[1]{\smallbreak\noindent\textbf{#1}}

\newcommand{\gdino}{Grounding DINO\xspace}
\newcommand{\methname}{ABRA\xspace}

\makeatletter
\DeclareRobustCommand\onedot{\futurelet\@let@token\@onedot}
\def\@onedot{\ifx\@let@token.\else.\null\fi\xspace}

\def\eg{\emph{e.g}\onedot} 
\def\ie{\emph{i.e}\onedot}

\definecolor{algc1}{HTML}{f7d779}
\definecolor{algc2}{HTML}{9fc5fc}

\renewcommand{\algorithmiccomment}[1]{\bgroup\hfill $\triangleright$ ~#1\egroup}

\definecolor{cvprblue}{rgb}{0.21,0.49,0.74}
\definecolor{customyellow}{HTML}{FFE629}
\definecolor{customblue}{HTML}{258CCC}

\def\methname{ABRA}
\def\methcomplete{Aligned Basis Relocation for Adaptation}

\usepackage{graphicx}
\newcommand{\fireflaticon}{{\includegraphics[height=1.25em]{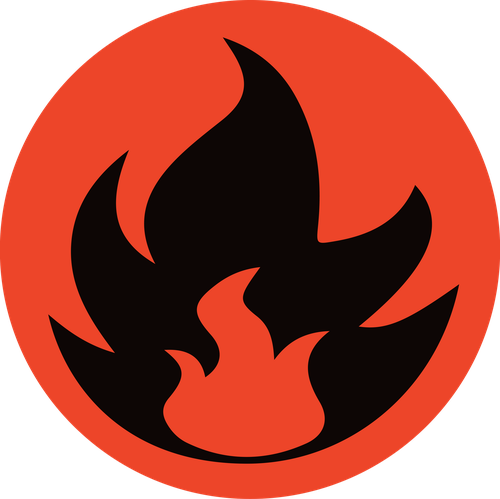}}}
\newcommand{\iceflaticon}{{\includegraphics[height=1.25em]{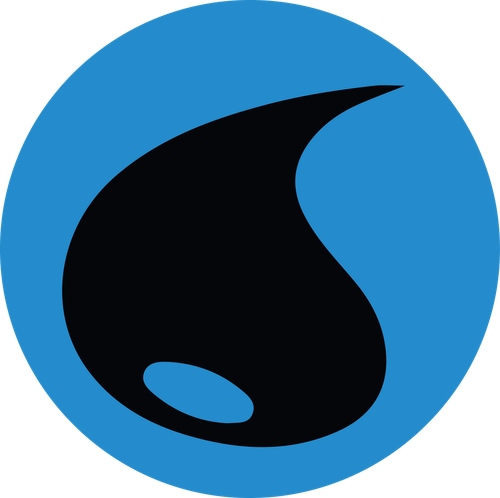}}}

\usepackage{placeins}

\makeatletter
\def\abra@prefix@section{Appendix~}
\newcommand{\appendixnumbering}{%
  \def\@seccntformat##1{\ifcsname abra@prefix@##1\endcsname\csname abra@prefix@##1\endcsname\fi\csname the##1\endcsname\@seccntDot\hskip 0.5em}%
  \crefalias{section}{appendix}%
  \counterwithin{table}{section}%
  \counterwithin{equation}{section}%
}
\makeatother

\ExplSyntaxOn
\cs_set:Npn \__first_footerline:
  {
    \group_begin:
    \small
    \sffamily
    \ifnum\theblind>0\relax
    \else
      \__short_authors:
    \fi
    \group_end:
  }
\ExplSyntaxOff

\begin{document}
\let\WriteBookmarks\relax
\def\floatpagepagefraction{1}
\def\textpagefraction{.001}
\shorttitle{ABRA: Teleporting Knowledge for Open-Vocabulary Detection}
\shortauthors{M.~Bernardi et~al.}

\title [mode = title]{ABRA: Teleporting
Fine-Tuned Knowledge Across Domains for Open-Vocabulary
Object Detection}

\author[1]{Mattia Bernardi}[
    orcid=0009-0003-3425-4148
]
\ead{mattia.bernardi@unimore.it}
\cormark[1]
\credit{Writing – original draft, Software, Methodology, Data curation, Conceptualization}

\author[1]{Chiara Cappellino}[
    orcid=0009-0002-5333-0524
]
\ead{chiara.cappellino@unimore.it}
\credit{Writing – original draft, Methodology, Data curation, Conceptualization}

\author[1]{Matteo Mosconi}[
    orcid=0009-0008-1989-5779
]
\ead{matteo.mosconi@unimore.it}
\credit{Writing – original draft, Methodology, Conceptualization}

\author[2]{Samuele Bertaina}[
    orcid=0009-0004-9201-8781
]
\ead{samuele.bertaina@intesasanpaolo.com}
\credit{Visualization, Validation}

\author[1]{Enver Sangineto}[
    orcid=0000-0002-5187-4133
]
\ead{enver.sangineto@unimore.it}
\credit{Writing – original draft, Methodology, Conceptualization}

\author[1]{Angelo Porrello}[
    orcid=0000-0002-9022-8484
]
\ead{angelo.porrello@unimore.it}
\credit{Supervision, Project administration, Methodology, Conceptualization}

\author[1]{Simone Calderara}[
    orcid=0000-0001-9056-1538
]
\ead{simone.calderara@unimore.it}
\credit{Supervision, Project administration, Methodology, Conceptualization}

\cortext[cor1]{Corresponding author}

\nonumnote{Published in Pattern Recognition Letters 211 (2027) 269--280, \url{https://doi.org/10.1016/j.patrec.2026.11.020}. \copyright\ 2026 The Author(s). CC BY-NC-ND 4.0.}

\affiliation[1]{
    organization={University of Modena and Reggio Emilia},
    city={Modena},
    country={Italy}
}
\affiliation[2]{
    organization={Intesa Sanpaolo},
    city={Turin},
    country={Italy}
}

\begin{abstract}
Although recent Open-Vocabulary Object Detection architectures, such as \gdino, demonstrate strong zero-shot capabilities, their performance degrades significantly under domain shifts. Moreover, many domains of practical interest, such as nighttime or foggy scenes, lack large annotated datasets, preventing direct fine-tuning. In this paper, we introduce \textbf{Aligned Basis Relocation for Adaptation (\methname)}, a method that transfers class-specific detection knowledge from a labeled source domain to a target domain where these classes have no annotations, and no images are selected for them. \methname\ decouples a detector into class-agnostic domain experts, obtained through Objectification, and class-specific residuals learned on top of them, and projects each source residual onto the singular subspaces of the target domain expert, without training. Extensive experiments across challenging domain shifts demonstrate that \methname\ successfully teleports class-level specialization under multiple adverse conditions. Our code is available on \href{https://github.com/brunzibeer/ABRA}{GitHub}.
\end{abstract}


\maketitle

\section{Introduction}
\label{sec:intro}
Recent advances in Open-Vocabulary Object Detection (OVD) demonstrate that Vision-Language Models~\cite{radford2021learning} can localize a broad array of concepts specified at inference time via natural-language prompts, effectively removing the constraint of fixed training categories. Despite this progress, performance often degrades sharply under domain shifts, such as transitions from daylight to low-light scenes or from clear weather to fog. Furthermore, in such adverse scenarios, acquiring large, fully annotated detection datasets proves prohibitively expensive or impractical.

Domain Adaptive Object Detection (DAOD) methods typically address this by training a detector on a labeled \emph{source} domain and generating pseudo-bounding boxes on unlabeled \emph{target} images for fine-tuning~\cite{tang2025sourcefreedomainadaptiveobject,10466767}. However, under severe domain shifts (\eg, nighttime imagery), these pseudo-labels are often unreliable. Moreover, requiring target-domain images containing all classes of interest -- even without bounding-box annotations -- implicitly introduces a weak form of supervision, as images must be preselected to ensure full class coverage~\cite{Sangineto_2019}.

In this paper, we propose a distinct adaptation scenario where only a subset of target-domain classes is available during training (\eg, \emph{car}, \emph{bicycle}), while other target classes (\eg, \emph{bus}, \emph{truck}) are labeled exclusively in the source domain, with \textbf{no bounding boxes}, and no images selected for them, in the target domain. This setting mirrors a common real-world problem where annotated data is readily available for frequent classes but absent for rare ones. Since no target images are selected for these classes, standard DAOD approaches are inapplicable because they usually need to generate pseudo-bounding boxes on training images containing the categories of interest. Formally, assuming a supervised \emph{source} domain, we partition \emph{target}-domain classes into \emph{available} and \emph{unavailable} sets, aiming to transfer knowledge specifically to the latter.

To address this, we build upon an OVD backbone (\ie, \gdino~\cite{liu2024grounding}) and decompose the adaptation process into two complementary levels of expertise: \emph{domain} and \emph{class}. At the domain level, we introduce \emph{objectification}, constructing a class-agnostic \emph{domain expert} to capture characteristics like illumination and texture. At the class level, we adapt \cite{lingam2024svft} to derive lightweight \emph{class experts} encoding category-specific semantics.

Building on this modular structure, we propose \textbf{\methname} (\textbf{Aligned Basis Relocation for Adaptation}), a method transferring class-level knowledge to a target domain where those classes have no annotations. \methname\ learns each class residual on top of the source domain expert and transports it onto the target domain expert by projecting it onto the singular subspaces of the latter, which reduces to plain addition on square layers. Unlike prior weight-space transfer methods (\cref{sec:related}), \methname\ transfers a class update across a domain shift.

We evaluate \methname\ on the Cityscapes~\cite{Cordts2016Cityscapes} $\rightarrow$ Foggy Cityscapes~\cite{SDV18} transfer, alongside four domain shifts within the Single-Source Domain Generalized Object Detection (SDGOD) setting~\cite{wu2022single}. Compared against several weight-space transfer and domain adaptation baselines, \methname\ consistently demonstrates superior performance, exhibiting strong zero- and few-shot transfer capabilities and providing improved weight initialization for downstream adaptation.
In summary, our contributions are:
\begin{itemize}
    \item A novel framework, \textbf{\methname}, that decouples a detector into class-agnostic domain experts, obtained through Objectification, and class-specific residuals learned on top of them, enabling the training-free transport of class knowledge toward target classes without annotations, and showing that residuals transfer only when learned on the source domain expert.
    \item An extensive validation across multiple domain shifts, demonstrating superior zero- and few-shot transfer robustness compared to existing baselines.
\end{itemize}

\section{Related Work}
\label{sec:related}
\tit{Open-Vocabulary Object Detection.}
Open-Vocabulary Object Detection (OVD) aligns vision and language features to detect arbitrary classes~\cite{gu2022open,zang2022open,zhong2022regionclip,zhou2022detecting,minderer2022simple,radford2021learning}, often leveraging massive pretraining~\cite{li2021grounded,liu2024grounding}. While some methods adapt models like \gdino~\cite{liu2024grounding} for Incremental Learning~\cite{deng2024zero,cappellino2025dithub,van2022three}, they do not explicitly address the cross-domain detection of identical classes. Our paper fills this gap by decoupling domain and class semantics to facilitate cross-domain transport.

\tit{Modular Deep Learning and Parameter-Efficient Fine-Tuning.}~Modular Deep Learning~\cite{pfeiffer2023modular} and Parameter-Efficient Fine-Tuning (PEFT), such as LoRA~\cite{hu2022lora} and its variants~\cite{zhang2023adalora,kopiczko2024vera,li2024vb,liu2024dora,lingam2024svft,meng2024pissa}, enable selective and efficient network adaptation. For example, SVFT~\cite{lingam2024svft} applies low-rank residuals within the SVD subspace. We embrace this modular philosophy, but uniquely separate adaptation into a large domain-level module and a compact, class-specific module, whereas standard PEFT of a target class requires its target annotations.

\tit{Domain Adaptive Object Detection.}
Most DAOD methods rely on Mean Teacher~\cite{10.5555/3294771.3294885} architectures and pseudo-labeling~\cite{tang2025sourcefreedomainadaptiveobject}. To mitigate unreliable pseudo-labels under strong domain shifts, recent works utilize adaptive data augmentation and uncertainty modeling~\cite{kennerley2024catexploitinginterclassdynamics,zhang2018mixupempiricalriskminimization,mattolin2022confmixunsuperviseddomainadaptation,10466767}. Other efforts explore Single-domain Generalizable Open-Set Object Detection~\cite{Yuan_2025_ICCV} to group completely unseen target objects into an ``unknown'' category. Conversely, our approach tackles a fundamentally different setting: transferring fine-tuned knowledge for specific, known classes that have no labels in the target environment, and for which no target images are selected. Since DAOD and source-free methods~\cite{tang2025sourcefreedomainadaptiveobject,10466767} align features or pseudo-label target images containing the classes of interest, they cannot be applied here.

\tit{Weight-Space Transfer.}
Prior works transport learned knowledge via model stitching~\cite{bansal2021revisiting,hernandez2022model,pan2023stitchable}, task arithmetic~\cite{ilharco2023editing,ortiz2023task}, or model rebasin~\cite{ainsworth2022git,rinaldi2025update,stoica2024zipit}. Task arithmetic~\cite{ilharco2023editing} combines task vectors within a single shared model, while Param$\Delta$~\cite{cao2025param} and task-vector re-basining~\cite{rinaldi2025update} transfer a weight update to a new release of the same base model, for the same task and data distribution. None of these methods addresses a domain shift. In contrast, we transfer a class update to a target domain where the class has no annotations, and learn it on an objectified source expert, which is what makes the transfer effective.

\section{Preliminaries}
\label{sec:background}
\begin{figure}
    \centering
    \includegraphics[width=0.92\linewidth]{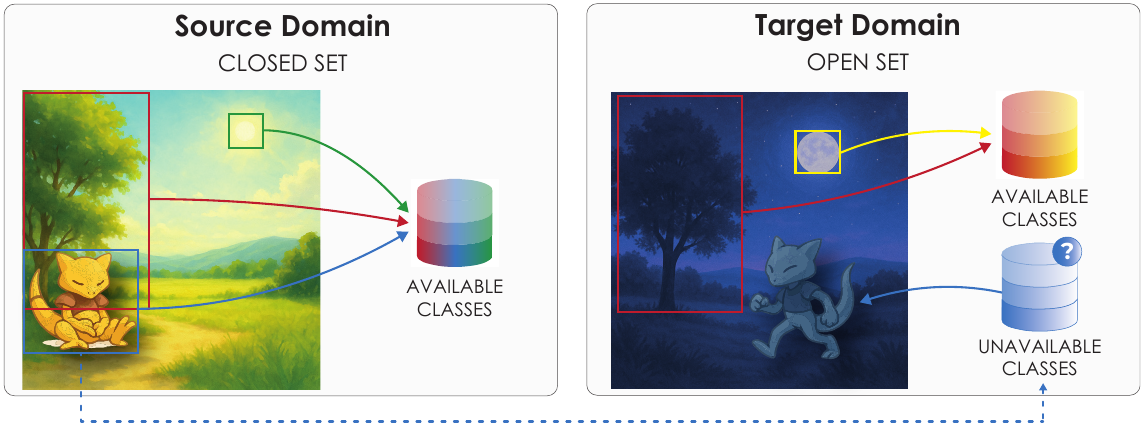}
    \caption{\textbf{Goal.} Transfer class-specific knowledge from the source domain to unavailable target classes without target-domain annotations of these classes.}
    \label{fig:setting}
    \vspace{-1.5em}
\end{figure}

\noindent\textbf{Problem Setting.}~Within our framework, we employ a pretrained Open-Vocabulary detector such as Grounding DINO~\cite{liu2024grounding} parameterized by $\theta_0$. As illustrated in~\cref{fig:setting}, our objective is to transfer class-specific knowledge from a \emph{source} domain to a \emph{target} domain. In the standard fully supervised setting, each domain is represented by a dataset of annotated images: $\mathcal{D}_S = \{(x_i^S, y_i^S)\}_{i=1}^{N_S}$ for the source domain and $\mathcal{D}_T = \{(x_i^T, y_i^T)\}_{i=1}^{N_T}$ for the target domain, where $x_i$ denotes an image and $y_i$ its ground-truth object annotations (bounding boxes and class labels).

Formally, we denote by $\mathcal{C}_S$ and $\mathcal{C}_T$ the sets of object classes defined in the source and target domains, respectively. We then partition the target classes into two disjoint sets, \ie, an \emph{available} set $\mathcal{C}_T^{\text{avail}}$ and an \emph{unavailable} set $\mathcal{C}_T^{\text{unav}}$, with $\mathcal{C}_T = \mathcal{C}_T^{\text{avail}} \cup\, \mathcal{C}_T^{\text{unav}}$ and $\mathcal{C}_T^{\text{avail}} \cap\, \mathcal{C}_T^{\text{unav}} = \emptyset$. Classes within $\mathcal{C}_T^{\text{avail}}$ are supported by annotated target-domain images, $\mathcal{D}_T^{\text{avail}}$, whereas classes in $\mathcal{C}_T^{\text{unav}} \subseteq \mathcal{C}_S$ are annotated exclusively in the source domain: no target-domain labels are available for them, and no target images are selected to contain them. Target images collected for the available classes may still include unlabeled instances of unavailable classes, which are treated as background. This fundamentally differs from standard DAOD, where target images are selected to contain all classes of interest and only lack supervision (\cref{sec:intro}).

\cref{fig:setting} exemplifies this setting with $\mathcal{C}_T^{\text{avail}} = \{ \emph{tree}, \emph{moon} \}$ and $\mathcal{C}_T^{\text{unav}} = \{ \emph{person}\}$. We emphasize that the classes targeted for transfer have no target-domain annotations.

\tit{Singular Vectors guided Fine-Tuning.}~To efficiently specialize our pretrained detector for novel classes, we leverage Singular Vectors guided Fine-Tuning (SVFT)~\cite{lingam2024svft}, a parameter-efficient fine-tuning strategy. Rather than updating full weight matrices, SVFT operates within the spectral domain, learning concise residuals $\Delta\Sigma$ on the singular values of a pretrained matrix $\theta_0 = U \Sigma V^\top$.  This yields the adapted parameterization $\theta_{FT} = U (\Sigma + \Delta\Sigma) V^\top$. This compact representation facilitates the creation of lightweight \emph{class modules} that are efficiently transportable across domains.

\section{\methcomplete{}}
\label{sec:method}
\begin{figure*}[t]
    \centering
    \includegraphics[width=\linewidth]{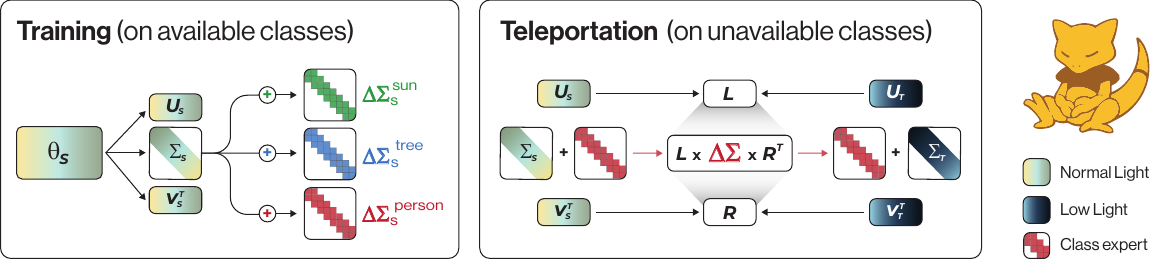}
    \caption{\textbf{Overview of the transfer pipeline.}
    Starting from domain experts obtained through Objectification,
    class-specific residuals learned via SVFT in the source domain
    are projected onto the target domain expert.}
    \label{fig:overview}
    \vspace{-1em}
\end{figure*}

We introduce \methname, a novel approach designed to transfer class-specific knowledge from a source domain to a target one, achieving this transition without any target-domain annotation of the transferred classes. We first use class-agnostic fine-tuning to create the source ($\theta_S$) and the target ($\theta_T$) domain modules. We then transport class-specific residuals from the source to the target module. The key design choice is this split: a residual learned on the source domain expert transfers to the target domain expert, whereas one learned on the pretrained model does not (\cref{tab:transfer_origin}). Specifically, we derive a class-specific expert for target class $c$, denoted as $\theta_{S}^{(c)} = \theta_{S} + \tau_{S}^{(c)}$, where $\tau_{S}^{(c)}$ represents the residual obtained by fine-tuning solely on class $c$ on source domain data. Our ultimate objective is to transfer this fine-tuning displacement, $\tau_{S}^{(c)} = \theta_{S}^{(c)} - \theta_{S}$, onto the target-domain module $\theta_T$, thereby yielding an estimate, $\hat{\theta}_{T}^{(c)}$, of the ideal detector $\theta_{T}^{(c)}$. \cref{fig:overview} illustrates the full method.

\tit{Organization.} In \cref{sec:objectification}, we detail the derivation of the domain expert models, $\theta_S$ and $\theta_T$. Additionally, \cref{sec:specialization} shows how we derive the class expert in the source domain: $\theta_{S}^{(c)} = \theta_{S} + \tau_{S}^{(c)}$. Finally, the teleporting process realized by \methname\ can be summarized as follows:
\begin{equation}
\hat{\theta}_T^{(c)} = \theta_T + \pi_{S \rightarrow T}\big(\tau_S^{(c)}\big), \quad \text{where } \tau_S^{(c)} = \theta_S^{(c)} - \theta_S
\label{eq:\methname_transfer}
\end{equation}
where $\pi_{S \rightarrow T}(\cdot)$ denotes the transport function that binds the source domain $S$ to the target domain $T$ (\cref{sec:rotation}).

\begin{figure}
    \centering
    \includegraphics[width=0.8\linewidth]{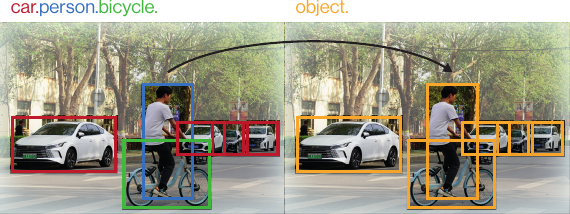}
    \caption{\textbf{Objectification.} Each domain is fine-tuned using a unified \textit{``object''} label to capture domain appearance.}
    \label{fig:objectification}
    \vspace{-1.5em}
\end{figure}

\subsection{Domain Expert through Objectification}
\label{sec:objectification}
In the upper-left section of \cref{fig:overview}, we illustrate the derivation of the domain expert $\theta_S$. We define a domain expert as a detection model that captures the underlying visual statistics of a specific domain (\eg, foggy or rainy weather conditions) while remaining strictly agnostic to individual classes.
To train this domain expert, we introduce a procedure termed \textit{\textbf{objectification}}, depicted in \cref{fig:objectification}. Specifically,  for each ground-truth bounding box belonging to the \emph{top three most represented} classes, the original semantic label (\eg, \texttt{car}, \texttt{person}, \texttt{bicycle}) is replaced with a single, neutral \texttt{``object''} category. This strategy compels the model to locate generic objects of interest rather than identifying their precise classes. These three classes are selected once, by source-domain frequency, and objectified in both domains; as they belong to $\mathcal{C}_T^{\text{avail}}$, no unavailable class is ever objectified. They cover $93.2\%$ of the source training boxes on Cityscapes (a fourth class adds $3.4\%$), $96.5\%$ on SDGOD Day-Clear, and $86$--$98\%$ of the boxes of every target training set.
Formally, given a specific domain, we construct an objectified dataset $\tilde{\mathcal{D}}_S$ by relabeling the bounding boxes of the top three represented classes as \texttt{``object''} while discarding all other annotations. Precisely:
\begin{equation}
\theta_S = \text{FT}(\theta_0, \tilde{\mathcal{D}}_S)\quad\text{where } \tilde{\mathcal{D}}_S = \big\{(x_i, \texttt{``object''})\big\}_{i=1}^{N_S^{\texttt{top-3}}}
\label{eq:domain_expert}
\end{equation}
where FT is a standard fine-tuning, $\theta_0$ denotes the pretrained parameters of \gdino, $x_i$ represents the input image, and $N_S^{\texttt{top-3}}$ is the total number of images containing at least one of the three dominant classes.
The resulting model, $\theta_S$, serves as the \textit{source domain expert}. An identical procedure is applied to derive the target domain expert, $\theta_T$, on all target training images containing at least one available class; these images are not filtered, so unlabeled instances of unavailable classes may appear, and are treated as background. The procedure can be generalized for any domain of interest.
\subsection{Class Experts through Singular Vectors}
\label{sec:specialization}
Let $c_S = \texttt{bus}$ denote a specific class that is supervised in the source dataset but has no annotations in the target domain. Building upon the domain expert $\theta_S$, we aim to train a lightweight, class-specific module specialized in detecting instances of this category.
To achieve this, we first filter the training set by retaining only images containing at least one instance of the class of interest, and strictly masking out all bounding boxes belonging to any other class. Subsequently, using the SVFT technique shown in \cref{sec:background}, we efficiently fine-tune a compact residual on top of the domain expert $\theta_S$. Specifically, for a source domain expert weight matrix $\theta_{S, \ell}$ at layer $\ell$, we first compute its Singular Value Decomposition, yielding $U_{S, \ell} \Sigma_{S, \ell} V_{S, \ell}^\top$. We then initialize a trainable, class-specific residual matrix $\Delta\Sigma_{S, \ell}^{(c)}$ while freezing the original components $U_{S, \ell}$, $\Sigma_{S, \ell}$, and $V_{S, \ell}^\top$. Thus, the forward pass $f$ for the $\ell$-th layer is  formulated as:
\begin{equation}
f_\ell(x) = \underbrace{U_{S,\ell}\vphantom{\Delta\Sigma_{S,\ell}^{(c)}}}_{\text{\scriptsize \raisebox{-0.15em}{\iceflaticon}\; frozen}} \Bigl( \underbrace{\Sigma_{S,\ell}\vphantom{\Delta\Sigma_{S,\ell}^{(c)}}}_{\text{\scriptsize \raisebox{-0.15em}{\iceflaticon}\; frozen}} + \underbrace{\Delta\Sigma_{S,\ell}^{(c)}}_{\text{\scriptsize \raisebox{-0.15em}{\fireflaticon}\; learnable}} \Bigr) \underbrace{V_{S,\ell}^\top\vphantom{\Delta\Sigma_{S,\ell}^{(c)}}}_{\text{\scriptsize \raisebox{-0.15em}{\iceflaticon}\; frozen}} x.
\label{eq:svft_learnable_pifont}
\end{equation}
This training paradigm naturally imposes a low-rank constraint on the residual matrix $\Delta\Sigma_{S,\ell}^{(c)}$. In practice, these residuals can be restricted to a diagonal (parameterized as a vector) or banded (\eg, tri-diagonal) structure~\cite{lingam2024svft}.

Upon concluding this fine-tuning phase, we obtain a class-specialized model $\theta_{S}^{(c)} = \theta_{S} + \tau_{S}^{(c)}$ tailored to class $c$ within domain $S$, as illustrated in the upper-right section of \cref{fig:overview}.
\subsection{Teleportation through Subspace Projection}
\label{sec:rotation}
Equipped with class-agnostic domain experts ($\theta_S$ and $\theta_T$) and class-specific residuals $\tau_{S}^{(c)}$, we now address adaptation to a target domain where class $c$ has no annotations, so that its target residual is unknown and must be inferred from the source one.

\tit{SVD form.} For clarity, we omit the layer index $\ell$. For each adapted weight matrix $W \in \mathbb{R}^{m \times n}$, we employ the thin SVD at full rank, without truncation: $U \in \mathbb{R}^{m \times k}$, $\Sigma \in \mathbb{R}^{k \times k}$, and $V \in \mathbb{R}^{n \times k}$, with $k = \min(m, n)$. Of the $66$ adapted layers, $24$ are square ($256 \times 256$) and $42$ are rectangular ($128 \times 256$, $1024 \times 256$, and $256 \times 1024$).

\tit{Transport as projection.} The main idea is to express the source residual $\tau_S^{(c)} = U_S \Delta\Sigma_S^{(c)} V_S^\top$ in the singular basis of the target domain expert. Its coefficients in this basis are:
\begin{equation}
C^{(c)} = U_T^\top \tau_S^{(c)} V_T = L\,\Delta\Sigma_S^{(c)} R^\top,
\label{eq:coefficients}
\end{equation}
with $L = U_T^\top U_S$ and $R = V_T^\top V_S$, which defines the transport function in \cref{eq:\methname_transfer} as $\pi_{S \rightarrow T}(\tau_S^{(c)}) = U_T C^{(c)} V_T^\top$. The target class expert then reads:
\begin{align}
\hat{\theta}_T^{(c)} &= U_T\bigl(\Sigma_T + C^{(c)}\bigr)V_T^\top = \theta_T + P_U\,\tau_S^{(c)}\,P_V, \label{eq:projection}\\
P_U &= U_T U_T^\top, \qquad P_V = V_T V_T^\top. \nonumber
\end{align}
In weight space, the transported residual is therefore the projection of the source residual onto the column and row spaces of $\theta_T$. On square layers, $U_T$ and $V_T$ are orthogonal, $P_U$ and $P_V$ reduce to the identity, and the transported residual coincides with $\tau_S^{(c)}$ up to floating-point precision. On rectangular layers, one of the two projections is non-trivial and alters the residual by $1.8$--$7.1\%$ (\cref{sup:optim}). Crucially, \methname\ thus effectively adds the source residual to the target domain expert: what makes this residual transferable is not the transport itself, but the model on which it is learned, \ie, the objectified source domain expert $\theta_S$ (\cref{tab:transfer_origin}). This procedure is illustrated in the lower section of \cref{fig:overview}.

\section{Experiments}
\label{sec:experiments}
\subsection{Benchmarking}
\label{sec:benchmarking}
\tit{Datasets.} As our setting introduces a novel domain transfer paradigm, no standardized benchmark exists, to the best of our knowledge. Our primary benchmark focuses on the widely adopted domain shift from Cityscapes~\cite{Cordts2016Cityscapes} to Foggy Cityscapes~\cite{SDV18}. The original Cityscapes dataset provides high-resolution urban street scenes captured under clear weather, while Foggy Cityscapes (0.02 intensity) offers the corresponding scenes rendered with synthetic fog. Here, $\mathcal{C}_S$ contains $8$ classes: \emph{car}, \emph{person}, and \emph{bicycle} are available in the target domain, while \emph{bus}, \emph{motorcycle}, \emph{rider}, \emph{train}, and \emph{truck} are unavailable. To further validate our approach across diverse environmental variations, we employ the SDGOD benchmark~\cite{wu2022single}, which encompasses five distinct weather and lighting conditions; Day-Clear serves as the source domain for Day-Foggy, Dusk-Rainy, Night-Clear, and Night-Rainy. Its $7$ classes split into the available \emph{car}, \emph{person}, and \emph{truck}, and the unavailable \emph{bus}, \emph{bike}, \emph{motorcycle}, and \emph{rider}, over which SDGOD averages are computed.

\tit{Target-domain data.} As the training images of $\theta_T$ are not filtered (\cref{sec:objectification}), $56.0\%$ of them on Foggy Cityscapes ($11$--$61\%$ on SDGOD) also contain unlabeled instances of unavailable classes. Retraining $\theta_T$ without any such image changes \methname\ by only $0.70$ mAP on average, while $\theta_T$ alone improves on these classes from $6.78$ to $15.84$ mAP: such instances act as background, not as leaked supervision.

\tit{Evaluation protocol.} In our setting, class-level evaluation is performed strictly on the subset of images containing at least one ground-truth instance of the target class, with all other bounding boxes masked out. This per-class protocol isolates class-level transfer: misclassifications of other objects in the evaluated images are still false positives, whereas images without the class are excluded, which raises absolute AP. On all $500$ test images (standard protocol), absolute scores decrease (\eg, \methname: $40.52 \rightarrow 30.05$ mAP), but the ranking of the methods is unchanged (supplementary material). Performance is reported in terms of mAP and AP50; the former represents an average over 10 increasing Intersection over Union (IoU) thresholds, while the latter is computed at a fixed IoU threshold of 50\%. Class experts, including those of the Source baseline, are selected on the source test split of their class; FDA uses its last epoch.

\noindent To evaluate our approach, we benchmark against several baselines. DAOD and source-free methods, as well as PEFT of a target class, require target images or annotations of the class and cannot be applied in our setting (\cref{sec:related}).
\tit{Target fine-tuning (\textit{reference}).} With the same SVFT recipe as the Source baseline, we train a model for each class $c$ starting from $\theta_0$, on the target images containing $c$, with the boxes of co-occurring classes masked out; its checkpoint is selected on the target test split. As it uses target annotations of $c$, which our setting excludes, this model is a reference rather than an upper bound: \methname\ starts from $\theta_T$, trained on a much larger pool of target images, and can exceed it on some classes (\cref{sec:main_results}).
\tit{\gdino Zero shot (\textit{baseline}).} The frozen pretrained model $\theta_0$.
\begin{table*}[t]
\centering
\caption{\textbf{Main Results on Target Domains.} Comparison of mAP on unavailable categories (Cityscapes $\rightarrow$ Foggy Cityscapes) and across challenging condition splits (SDGOD), per-class protocol (\cref{sec:benchmarking}). More metrics are provided in the supplementary material. \methname\ outperforms all baselines on average. \textit{Target fine-tuning} and \methname-$\alpha$ use target labels of unavailable classes and are shown for reference; FDA uses unlabeled target images containing them. \textbf{Bold}: best excluding these two rows. $^\dagger$Depends on SVD signs.}
\label{tab:main_results_merged}
\begingroup
\setlength{\tabcolsep}{4pt}
\resizebox{\linewidth}{!}{
\begin{tabular}{l *{5}{S} | S || *{4}{S} | S}
\toprule
& \multicolumn{6}{c||}{\textbf{Cityscapes $\rightarrow$ Foggy Cityscapes}}
& \multicolumn{5}{c}{\textbf{SDGOD}} \\
\cmidrule(lr){2-7}
\cmidrule(lr){8-12}
\textbf{Method}
& {\textbf{Bus}} & {\textbf{Motor}} & {\textbf{Rider}} & {\textbf{Train}} & {\textbf{Truck}} & {\textbf{Average}}
& {\textbf{Day Foggy}} & {\textbf{Dusk Rainy}} & {\textbf{Night Clear}} & {\textbf{Night Rainy}} & {\textbf{Average}} \\
\midrule
\textit{Target fine-tuning (reference)}
& {58.76} & {31.23} & {43.88} & {33.06} & {39.92} & {41.37}
& {36.37} & {26.84} & {37.51} & {16.82} & {29.38} \\
\midrule
Zero shot
& {48.61} & {23.13} & {18.97} & {16.31} & {31.20} & {27.65}
& {26.36} & {19.55} & {27.50} & {9.19} & {20.65} \\
Source
& {54.12} & {28.95} & {\textbf{42.49}} & {33.40} & {36.78} & {39.15}
& {31.77} & {27.62} & {\textbf{36.38}} & {15.25} & {27.75} \\
FDA~\cite{yang2020fda}
& {51.53} & {25.12} & {40.07} & {30.92} & {33.97} & {36.32}
& {31.55} & {25.94} & {35.72} & {14.26} & {26.87} \\
Param$\Delta$~\cite{cao2025param}$^\dagger$
& {37.42} & {7.29} & {9.28} & {7.78} & {13.74} & {15.10}
& {17.69} & {5.17} & {4.85} & {7.83} & {8.88} \\
\midrule
\rowcolor{customyellow!55}
\textbf{ABRA (ours)}
& {\textbf{57.22}} & {\textbf{30.01}} & {42.29} & {\textbf{34.97}} & {\textbf{38.12}} & {\textbf{40.52}}
& {\textbf{32.34}} & {\textbf{27.90}} & {35.90} & {\textbf{16.12}} & {\textbf{28.06}} \\
\midrule
\textit{ABRA-$\alpha$ (oracle)}
& {57.23} & {30.19} & {42.29} & {38.53} & {38.60} & {41.37}
& {33.55} & {28.46} & {35.99} & {17.28} & {28.82} \\
\bottomrule
\end{tabular}
}
\endgroup
\vspace{-1em}
\end{table*}

\tit{Source (\textit{baseline}).} This baseline evaluates a model $\theta_{FT}^{(c)}$ directly on the target domain, where $\theta_{FT}^{(c)}$ is obtained by fine-tuning, via SVFT, the pretrained model $\theta_0$ on class $c$ within the source domain. This setting quantifies the performance degradation caused by directly transferring a source-trained model to a novel target domain.
\tit{FDA~\cite{yang2020fda} (\textit{competitor}).} Fourier Domain Adaptation transfers the low-frequency amplitude spectrum of target images onto source images. We fine-tune $\theta_0$ via SVFT on the source images of class $c$, stylized with unlabeled target training images that contain $c$. FDA thus uses strictly more target information than \methname, as our setting selects no target image for $c$.
\tit{Param$\Delta$~\cite{cao2025param} for Direct Weight Mixing (\textit{competitor}).}~Param$\Delta$ transfers fine-tuned knowledge across models without additional training by directly injecting task vectors. We adapt it to the SVFT parameterization by reusing the source coefficients $\Delta\Sigma_{S}^{(c)}$ as-is in the singular basis of the target expert, without projection:
\begin{equation}
    \widehat{\Delta\Sigma}_{T}^{(c)}
    = \Delta\Sigma_{S}^{(c)} , \Longrightarrow \theta_{T,\ell}^{(c)} \approx U_{T}\bigl(\Sigma_{T} + \Delta\Sigma_{S}^{(c)} \bigr) V_{T}^\top.
    \label{eq:reprojectionparamdelta}
\end{equation}
As singular vectors are defined only up to sign, these coefficients are not aligned with the target basis, and the result depends on the signs returned by the SVD. Adding the residual directly in weight space is analyzed in \cref{tab:transfer_origin}.
\tit{\methname-$\alpha$ (\textit{oracle}).} We scale the transported residual by $\alpha$, selected per class on the mAP of the target test split ($\alpha \in [0.1, 10)$, step $0.1$). As this uses target annotations, \methname-$\alpha$ is an oracle analysis, not a deployable variant; all our claims rely on plain \methname\ ($\alpha = 1$). On Cityscapes, the selected values are $1.1$ (Bus), $0.9$ (Motorcycle), $1.0$ (Rider), $1.4$ (Train), and $1.1$ (Truck).
\subsection{Main Results}
\label{sec:main_results}
\noindent \cref{tab:main_results_merged} summarizes our evaluation across target scenarios.
\tit{Cityscapes$\rightarrow$Foggy Cityscapes.} \textbf{\methname} proves highly effective across all transported classes. It reaches $40.52$ mAP, within $0.85$ of the Target fine-tuning reference without any target annotation of these classes, and outperforms Source ($39.15$), FDA ($36.32$), which uses target images of the class, and Param$\Delta$ ($15.10$). On Train, even untuned \methname\ exceeds the reference ($34.97$ vs.\ $33.06$): it starts from $\theta_T$, adapted to fog on all images with available classes, whereas the reference starts from $\theta_0$ and uses only the target instances of Train. With only $23$ test boxes, Train is also high-variance: under the standard protocol, the reference leads ($17.53$ vs.\ $13.53$).
\tit{SDGOD Splits.} Where baselines experience significant drops and Param$\Delta$~\cite{cao2025param} completely collapses, \textbf{\methname} achieves the highest overall average mAP, $28.06$ vs.\ $27.75$ for Source and $26.87$ for FDA. Though the \textit{Source} baseline marginally leads on the Night Clear split, our method dominates across all other conditions (Day Foggy, Dusk Rainy, Night Rainy) and consistently tracks the \textit{Target fine-tuning} reference.

\subsection{Additional Analysis}
\label{sec:additional}
\begin{figure}
    \centering
    \includegraphics[width=0.85\linewidth]{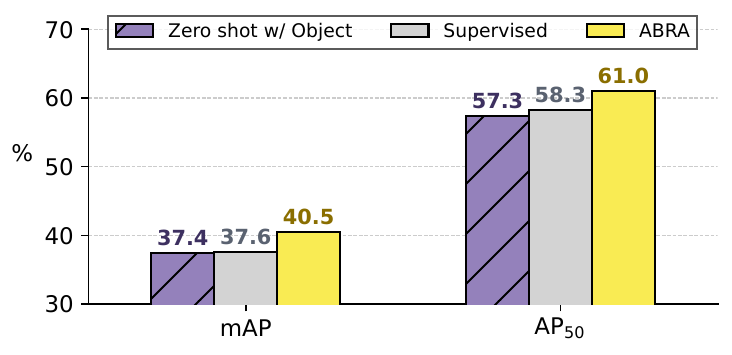}
    \caption{\textbf{Domain expert ablation.} Evaluating objectification and class visibility strategies. \methname\ yields the highest performance on unavailable classes.}
    \label{fig:domain_ablation}
    \vspace{-1.5em}
\end{figure}

\tit{Domain Design Choice.} To validate the design choices behind our domain experts, we conduct an ablation study comparing models trained via the proposed Objectification strategy against two alternatives. First, we establish an initial baseline (\textit{Zero Shot w/ Obj.}), in which the domain experts are trained with Objectification on the zero-shot detections of $\theta_0$ (score $\geq 0.3$, all Cityscapes categories), instead of the ground-truth boxes of the available classes. Second, we fine-tune the network using the actual semantic labels of the available categories---rather than a unified generic label (\textit{Supervised}). As demonstrated in~\cref{fig:domain_ablation}, both retaining the original ground-truth semantics and relying on zero-shot predictions inherently restrict transferability to \textit{unavailable} classes. Ultimately, \textit{Objectification} achieves the highest overall performance, confirming that isolating class-agnostic visual cues is essential for effective zero-shot cross-domain adaptation. Varying the number of objectified classes $k \in \{1, 2, 3\}$ changes the average of Bus and Train by less than $2$ mAP, with no $k$ best on both; we retain $k = 3$ for its box coverage (supplementary material).
\tit{\methname\ for Downstream Tasks.}~We use our transportable class-specific residuals to initialize two distinct strategies: supervised SVFT fine-tuning on target data of the class and Unsupervised Domain Adaptation, specifically adopting Fourier Domain Adaptation (FDA)~\cite{yang2020fda} for the latter. As detailed in~\cref{tab:initialization_comparison}, weights initialized via \methname\ consistently yield higher performance than the standard pretrained backbone $\theta_0$ across both methods. With FDA, \methname's initialization is mostly retained ($40.50$ vs.\ $40.52$ for \methname\ alone).
\begin{table}[t]
    \centering
    \caption{\textbf{Impact of initialization.} Initializing downstream tasks (SVFT fine-tuning or FDA~\cite{yang2020fda}) with \methname\ consistently outperforms $\theta_0$ pretraining.}
    \label{tab:initialization_comparison}
    \setlength{\tabcolsep}{4pt}
    \begin{tabular}{l c c c c}
        \toprule
        & \multicolumn{2}{c}{\textbf{SVFT}} & \multicolumn{2}{c}{\textbf{FDA}} \\
        \cmidrule(lr){2-3} \cmidrule(lr){4-5}
        \textbf{Init.} & \textbf{mAP} & \textbf{AP$_{50}$} & \textbf{mAP} & \textbf{AP$_{50}$} \\
        \midrule
        $\theta_0$ & 41.37 & 62.52 & 36.32 & 56.08 \\
        \rowcolor{customyellow!55}
        \methname\ & \textbf{42.80} & \textbf{62.81} & \textbf{40.50} & \textbf{61.24} \\
        \bottomrule
    \end{tabular}
\end{table}

\begin{table}[t]
\centering
\caption{\textbf{Where the transfer comes from.} Model on which the class residual $\tau$ is learned, and model it is added to. Cityscapes $\rightarrow$ Foggy Cityscapes, mAP on the unavailable classes, per-class protocol.}
\label{tab:transfer_origin}
\setlength{\tabcolsep}{2.5pt}
\resizebox{\linewidth}{!}{
\begin{tabular}{l c *{5}{S} S}
\toprule
\textbf{Model} & \textbf{$\tau$ on} & {\textbf{Bus}} & {\textbf{Motor}} & {\textbf{Rider}} & {\textbf{Train}} & {\textbf{Truck}} & {\textbf{Avg}} \\
\midrule
$\theta_T$ alone & -- & {22.72} & {3.31} & {5.85} & {2.17} & {6.26} & {8.06} \\
$\theta_0+\tau$ (Source) & $\theta_0$ & {54.12} & {28.95} & {42.49} & {33.40} & {36.78} & {39.15} \\
$\theta_T+\tau$ & $\theta_0$ & {45.61} & {12.47} & {34.56} & {21.26} & {28.10} & {28.40} \\
$\theta_T+\tau$ (direct) & $\theta_S$ & {57.27} & {30.09} & {42.36} & {35.29} & {38.32} & {40.66} \\
\rowcolor{customyellow!55}
\methname: $\theta_T+P_U\tau P_V$ & $\theta_S$ & {57.22} & {30.01} & {42.29} & {34.97} & {38.12} & {40.52} \\
\bottomrule
\end{tabular}
}
\end{table}

\tit{Where does the transfer come from?} We learn each class residual with the same recipe on either $\theta_S$ or $\theta_0$, and add it to the same $\theta_T$ (\cref{tab:transfer_origin}). Alone, $\theta_T$ scores $8.06$ mAP on the unavailable classes, so the gain comes from the residual. A residual learned on $\theta_0$ performs well on $\theta_0$ ($39.15$, \ie, Source) but drops to $28.40$ on $\theta_T$, whereas the residual learned on $\theta_S$ reaches $40.66$ when added directly and $40.52$ when projected, improving on every class. A class residual thus transfers when learned on a source expert aligned with $\theta_T$, which is precisely what Objectification provides.

\begin{figure}
    \centering
    \includegraphics[width=0.85\linewidth]{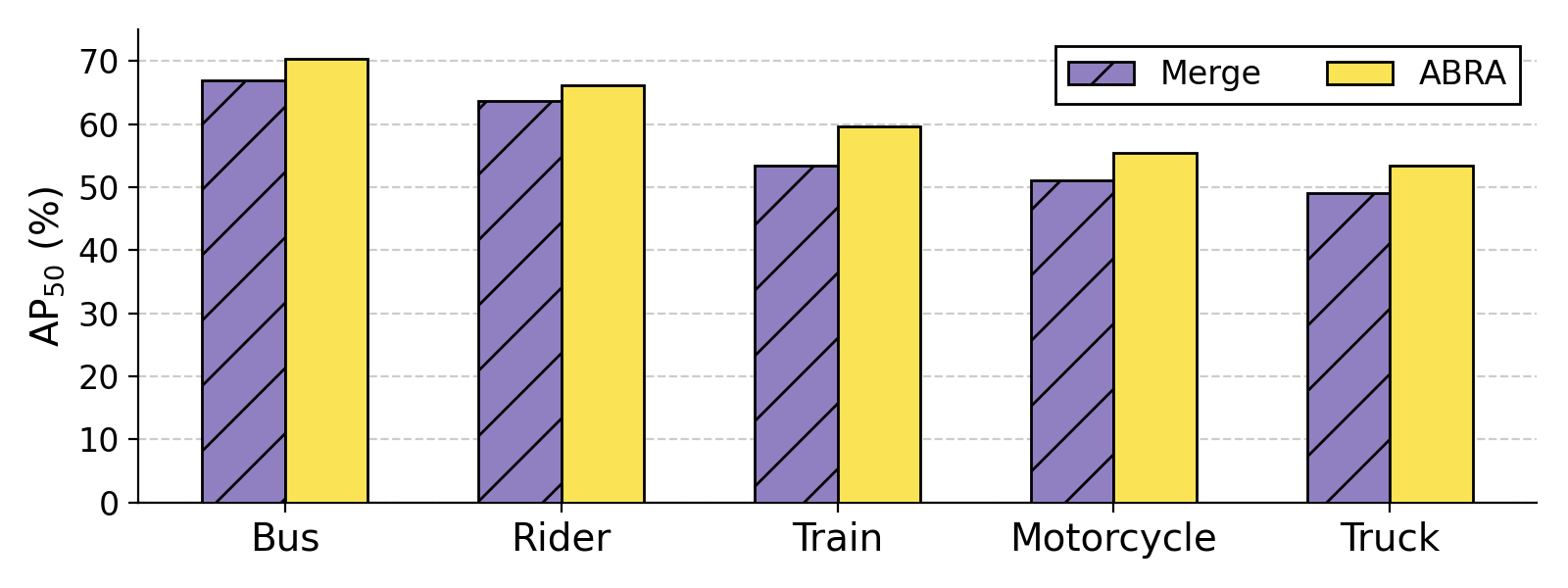}
    \caption{\textbf{Effect of specialization on Cityscapes.} Dedicating one expert per class (\methname) yields higher AP$_{50}$ than using a single unified expert (Merge).}
    \label{fig:specialization}
    \vspace{-1.5em}
\end{figure}

\tit{Effect of specialization.}~
We evaluate a baseline configuration where a single unified expert is trained to comprehend and transport all target classes simultaneously (denoted as ``Merge''). As shown in~\cref{fig:specialization}, dedicating a distinct, lightweight expert per class consistently yields higher AP$_{50}$ scores across all categories; the gap is largest for Train ($+6.2$ AP$_{50}$) and smallest for Rider ($+2.4$).

\begin{figure*}
    \centering
    \includegraphics[width=0.92\linewidth]{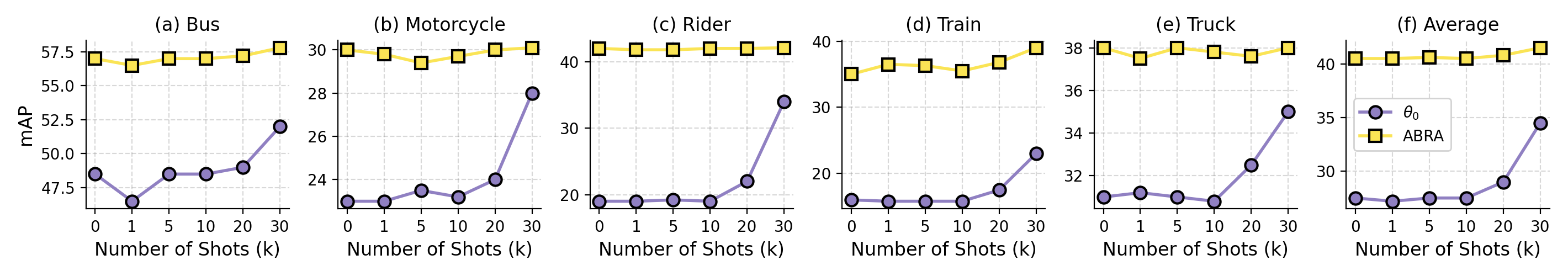}
     \caption{\textbf{Few-shot transfer evaluation.} Cityscapes comparison (mAP) between the standard pretrained backbone ($\theta_0$) and \methname\ across varying numbers of shots.}
    \label{fig:fewshot2}
    \vspace{-1.5em}
\end{figure*}

\tit{Few-Shot Transfer Evaluation.} We evaluate performance in the few-shot setting, starting from both the standard pretrained checkpoint ($\theta_{0}$) and our proposed approach. We focus on adapting to the target domain classes using 1, 5, 10, 20, and 30 shots. The results are presented in~\cref{fig:fewshot2}. As illustrated, \methname\ consistently outperforms the standard $\theta_{0}$ initialization across all few-shot scenarios and all evaluated categories.

\section{Conclusions}
\label{sec:conclusion}
We introduced \textbf{Aligned Basis Relocation for Adaptation (\methname)}, a novel domain knowledge transfer framework for the challenging scenario where the target classes of interest have no target-domain annotations. Building on \textit{Objectification} to derive domain experts, \methname\ leverages \textit{Specialization} to encode source-domain class knowledge as residuals learned on top of them, and projects these residuals onto the target domain expert, enabling the training-free relocation of this class-specific information to the target domain. Extensive experiments demonstrate \methname's strong, consistent transfer performance across diverse adverse shift conditions. On these benchmarks, source and target domain experts remain close enough in weight space that adding the source residual suffices. Under larger shifts, such as synthetic-to-real or RGB-to-thermal, the experts may drift apart and the residual may need to be adapted to the target expert; on Day-Foggy, mild masks of the residual already yield gains of up to $0.9$ mAP (supplementary material).

\printcredits

\section*{Declaration of competing interest}
The authors declare that they have no known competing financial interests or personal relationships that could have appeared to influence the work reported in this paper.

\section*{Data availability}
The repository is linked in the article.

\appendix
\appendixnumbering

\section{Transport as Subspace Projection}
\label{sup:optim}
For an adapted layer $W \in \mathbb{R}^{m \times n}$, let $\theta_S = U_S \Sigma_S V_S^\top$ and $\theta_T = U_T \Sigma_T V_T^\top$ be the thin SVDs at full rank of the source and target domain experts, with $U \in \mathbb{R}^{m \times k}$, $V \in \mathbb{R}^{n \times k}$, $k = \min(m, n)$, and $U^\top U = V^\top V = I_k$. The class residual learned on the source expert is $\tau_S = U_S \Delta\Sigma_S V_S^\top$, and \methname\ builds the target class expert as
\begin{equation}
\hat{\theta}_T = U_T\bigl(\Sigma_T + L\,\Delta\Sigma_S\,R^\top\bigr)V_T^\top,
\end{equation}
with $L = U_T^\top U_S$ and $R = V_T^\top V_S$.
Expanding the transported term gives
\begin{align}
U_T L\,\Delta\Sigma_S R^\top V_T^\top &= U_T U_T^\top \, U_S \Delta\Sigma_S V_S^\top \, V_T V_T^\top \nonumber\\
&= P_U\, \tau_S\, P_V,
\end{align}
where $P_U = U_T U_T^\top$ and $P_V = V_T V_T^\top$ are the orthogonal projectors onto the column and row spaces of $\theta_T$.

\tit{Square layers.} When $m = n$, $U_T$ and $V_T$ are orthogonal matrices, so $P_U = P_V = I$ and the transported residual equals $\tau_S$: \methname\ reduces to adding the source residual. Computed in float64, the relative change $\|P_U\tau_S P_V - \tau_S\|_F / \|\tau_S\|_F$ is $3.0$--$3.3\cdot10^{-15}$ on the $24$ square layers.

\tit{Rectangular layers.} When $m > n$, $V_T$ is square and $P_V = I$, while $P_U$ projects $\mathbb{R}^m$ onto the $k$-dimensional column space of $\theta_T$; when $m < n$, the roles are swapped. Exactly one side is therefore projected on every rectangular layer. On Cityscapes, the relative change is $1.8$--$7.1\%$ (median $3.1\%$) on $35$ of the $42$ rectangular layers. On the remaining seven, \ie, the six projections of the first fusion block and the attention-weight projection of the first encoder layer, it is about $5\cdot10^{-7}$, as source and target singular bases coincide up to float32 precision. On SDGOD Day-Foggy, it reaches $11.6\%$.

\tit{Nature of $L$ and $R$.} $L$ and $R$ are $k \times k$ matrices that are orthogonal only when source and target bases span the same subspace. On square layers, $\|L^\top L - I\|_F \approx 6\cdot10^{-4}$ only reflects the float32 storage of the source basis; on rectangular layers, $\|L^\top L - I\|_F$ reaches $0.25$ and $\|R^\top R - I\|_F$ reaches $0.28$. Hence, $L$ and $R$ are not rotations, and \methname\ does not solve an orthogonal alignment problem: it expresses the source residual in the singular basis of the target expert, which in weight space is the projection above.

\section{Implementation Details}
\methname\ and the compared baselines all adapt \gdino as the pretrained OVD backbone. Domain checkpoints are derived by fine-tuning only the encoder attention layers for $10$ epochs, utilizing a standard learning rate of $1\mathrm{e}{-4}$, a batch size $\mathcal{B}=2$, and a weight decay of $1\mathrm{e}{-4}$. Starting from the source-domain expert, we perform class specialization via SVFT by training these same attention layers for an additional $12$ epochs with an increased learning rate of $1\mathrm{e}{-2}$ and $\mathcal{B}=4$. For these class experts, we do not apply weight decay and utilize the banded pattern of SVFT with an off-diagonal value (\texttt{off\_diag}) of $2$. During both the domain and class expert training phases, we employ a multi-step learning rate scheduler that drops the learning rate by a factor of $10$ at epochs $7$ and $9$. Additionally, the specific layers \texttt{ref\_point\_head} and \texttt{sampling\_offset} are always assigned an initial learning rate reduced by a factor of $10$ with respect to the standard learning rate across both stages. The Source baseline and the Target fine-tuning reference follow the class-expert recipe, starting from $\theta_0$. All evaluations use batch size $1$, seed $0$, and the \gdino validation transforms; under the per-class protocol, the model is prompted with the name of the evaluated class.

\tit{FDA.} We fine-tune the encoder attention layers with SVFT (banded, \texttt{off\_diag} $= 2$), starting from $\theta_0$, with the FDA low-frequency ratio $L = 0.01$, for $12$ epochs with $\mathcal{B} = 4$, AdamW, a learning rate of $1\mathrm{e}{-2}$ ($1\mathrm{e}{-3}$ for \texttt{ref\_point\_head} and \texttt{sampling\_offset}), no weight decay, no scheduler, and seed $0$. Source images are those of the class, with the boxes of other classes removed; the style pool contains the target training images that contain the class, whose labels are not used. Images are resized to $1024 \times 512$, and an epoch has $\min(\#\text{source}, \#\text{style})$ batches, as source and style loaders are paired. We use the last epoch. When initialized with \methname\ (\cref{tab:initialization_comparison}), the learning rates are $1\mathrm{e}{-6}$ and $1\mathrm{e}{-7}$.

\tit{SVFT fine-tuning from \methname.} Starting from the transported model, we train the encoder attention layers with SVFT (banded, \texttt{off\_diag} $= 2$) on the target data of the class, for $12$ epochs with $\mathcal{B} = 4$, AdamW, a learning rate of $1\mathrm{e}{-2}$ ($1\mathrm{e}{-3}$ for \texttt{ref\_point\_head} and \texttt{sampling\_offset}), no weight decay, and the multi-step scheduler above.

\tit{Few-shot fine-tuning.} For \cref{fig:fewshot2}, both initializations are fine-tuned with SVFT (banded, \texttt{off\_diag} $= 2$) for $12$ epochs with $\mathcal{B} = 4$, on $k \in \{1, 5, 10, 20, 30\}$ target training images per class; the $k$-shot subsets are drawn once and shared by both initializations. The learning rate is $1\mathrm{e}{-4}$ for $k \leq 10$, $1\mathrm{e}{-3}$ for $k = 20$, and $5\mathrm{e}{-3}$ for $k = 30$, for both initializations. Each point is a single run.

\tit{Ablation models.} The \textit{Supervised} domain expert of \cref{fig:domain_ablation} is trained like the objectified one, but with the names of the three available classes instead of ``object''; for \textit{Zero Shot w/ Obj.}, both domain experts are trained with Objectification on the zero-shot detections of $\theta_0$. In both cases, class experts are then trained with the class-expert recipe on the corresponding source domain expert and transported with \methname. The \textit{Merge} expert of \cref{fig:specialization} is a single residual, trained with the class-expert recipe on the source images containing any of the five unavailable classes, prompted with their five names, and selected on the source test split of these classes.

\tit{Checkpoint selection.} Class experts, including those of the Source baseline, are selected on the source test split of their class. FDA uses its last epoch. The Target fine-tuning reference, the SVFT fine-tuning from \methname, and the few-shot runs use target annotations by design, and are selected on the target test split of their class.

\tit{Oracle $\alpha$ search.} For \methname-$\alpha$, we evaluate $\alpha \in [0.1, 10)$ with step $0.1$, stopping after $10$ consecutive decreases of mAP, and select the $\alpha$ with the highest mAP on the target test split of the class (on ties, the $\alpha$ closest to $1$). AP$_{50}$ and AP$_{75}$ are reported at that same $\alpha$. \cref{tab:alpha_values} lists the selected values.
\begin{table}[t]
\centering
\caption{\textbf{Selected $\alpha$ of \methname-$\alpha$ (oracle)}, per class and target split, chosen on mAP.}
\label{tab:alpha_values}
\setlength{\tabcolsep}{4pt}
\begin{tabular}{l c c c c c}
\toprule
\textbf{Cityscapes} & Bus & Motor & Rider & Train & Truck \\
Foggy Cityscapes & 1.1 & 0.9 & 1.0 & 1.4 & 1.1 \\
\midrule
\textbf{SDGOD} & Bus & Bike & Motor & Rider & \\
Day Foggy & 0.8 & 0.7 & 0.5 & 0.8 & \\
Dusk Rainy & 0.9 & 0.9 & 1.3 & 0.8 & \\
Night Clear & 0.9 & 1.0 & 1.1 & 1.0 & \\
Night Rainy & 0.5 & 1.1 & 0.9 & 0.9 & \\
\bottomrule
\end{tabular}
\end{table}

\section{Qualitative Results}

In this section, we provide additional qualitative results to further demonstrate the effectiveness of our proposed approach, \methname.
\cref{tab:qualitatives} illustrates a visual comparison between the source-only baseline, our method, and the ground truth (GT) on challenging examples from the target domain, which features severe fog and reduced visibility. As observed in the leftmost column, the baseline model struggles significantly under these adverse weather conditions: it misses the bus (top row) and the train (bottom row), and one of the two riders (middle row). Conversely, our proposed method (middle column) detects the bus and both riders, and places a loose box on the train. Predictions are shown at score $\geq 0.35$, and insets magnify the same region in the three panels of each row.
\begin{table*}
    \centering
    \begin{tabular}{@{}c@{\hspace{0.0125\linewidth}}c@{\hspace{0.0125\linewidth}}c@{}}
        Source (baseline) &
        \methname &
        GT \\
        \includegraphics[width=0.325\linewidth]{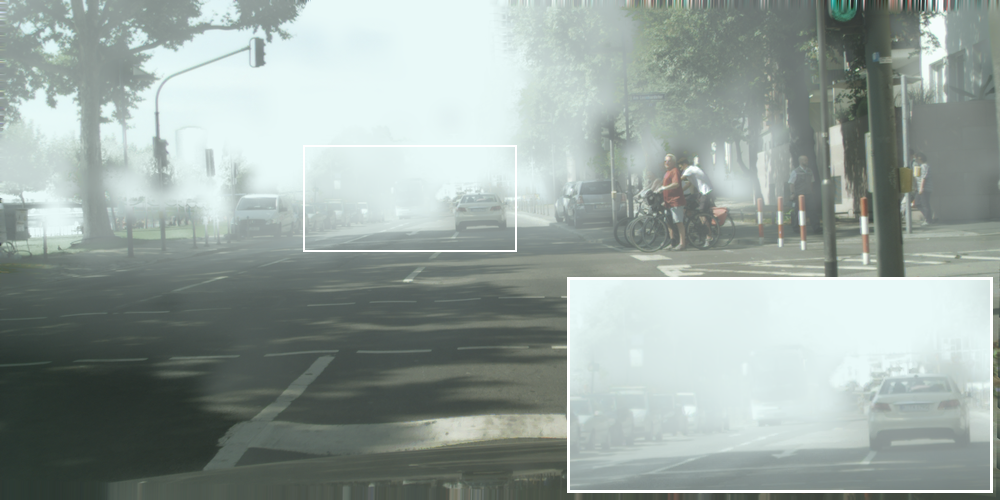} &
        \includegraphics[width=0.325\linewidth]{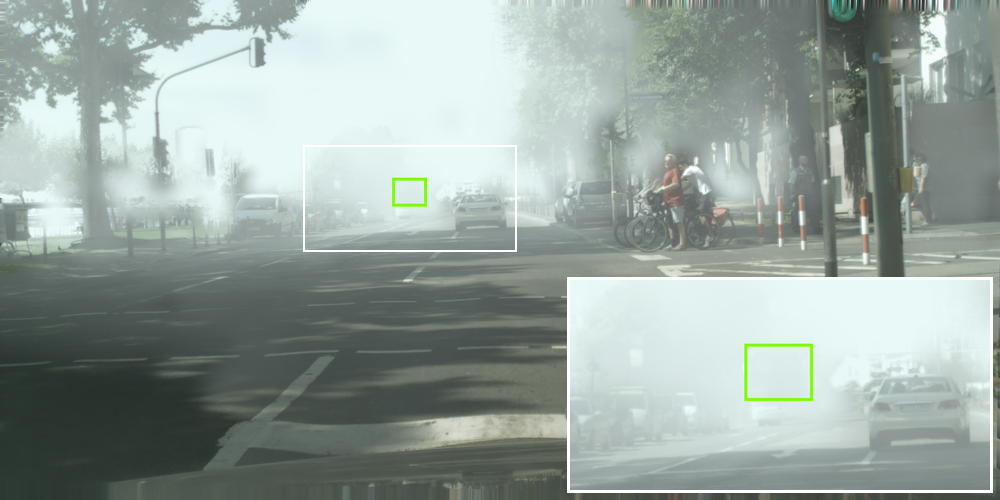} &
        \includegraphics[width=0.325\linewidth]{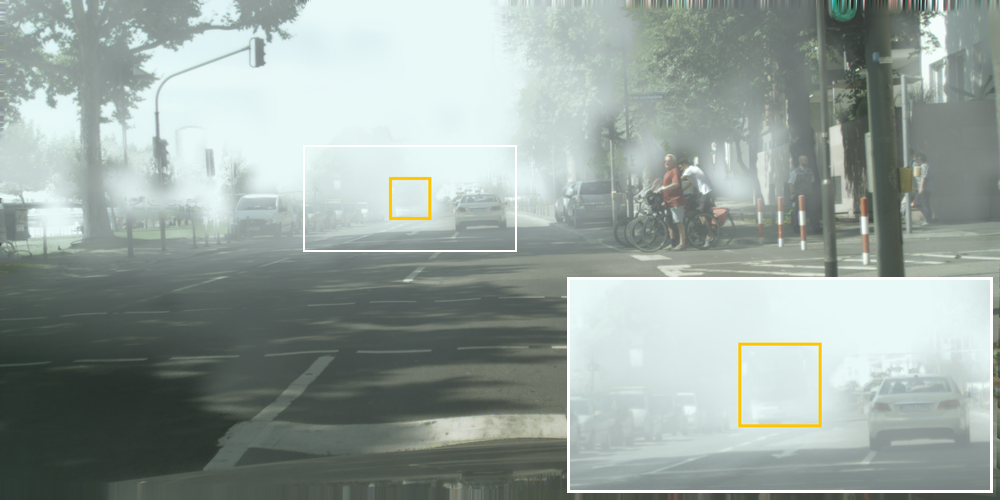} \\
        \includegraphics[width=0.325\linewidth]{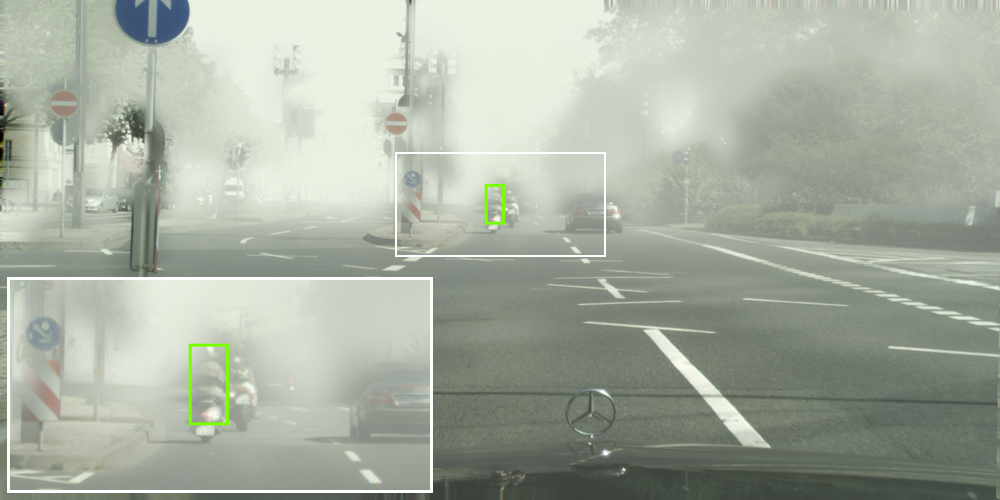} &
        \includegraphics[width=0.325\linewidth]{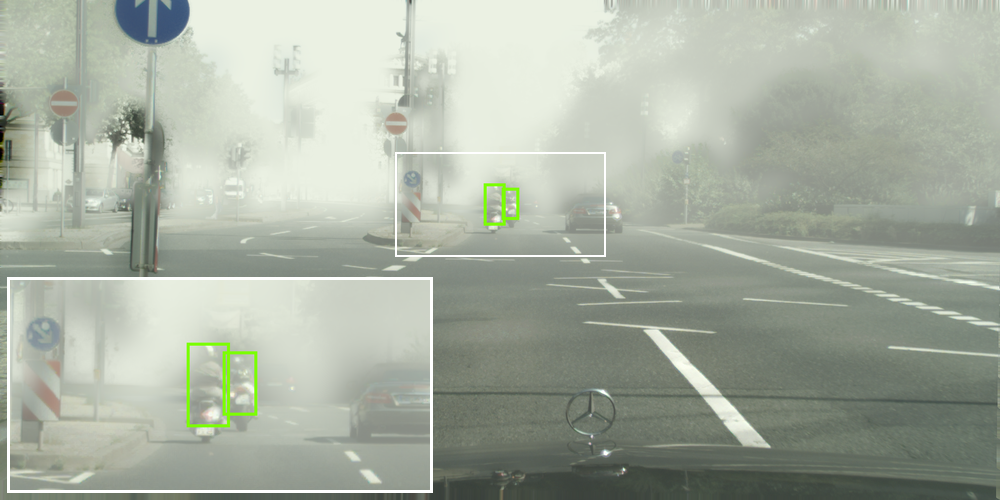} &
        \includegraphics[width=0.325\linewidth]{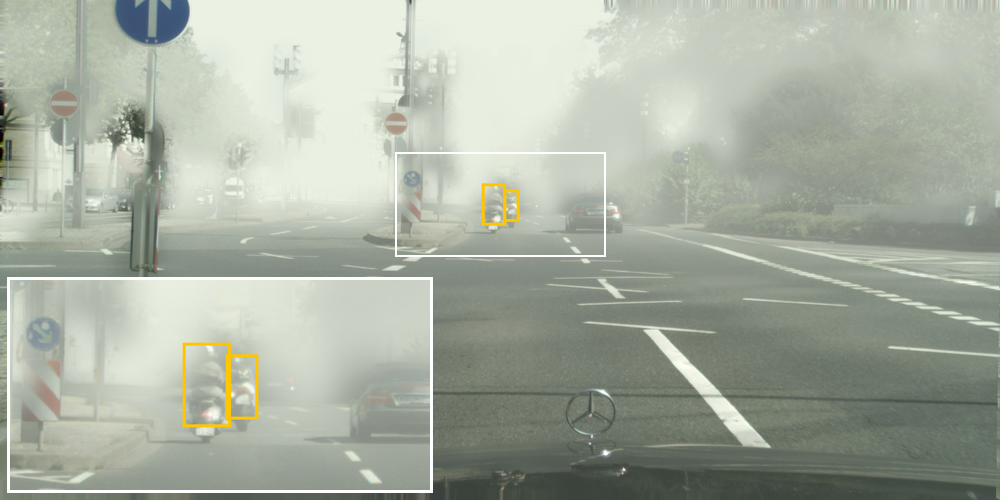} \\
        \includegraphics[width=0.325\linewidth]{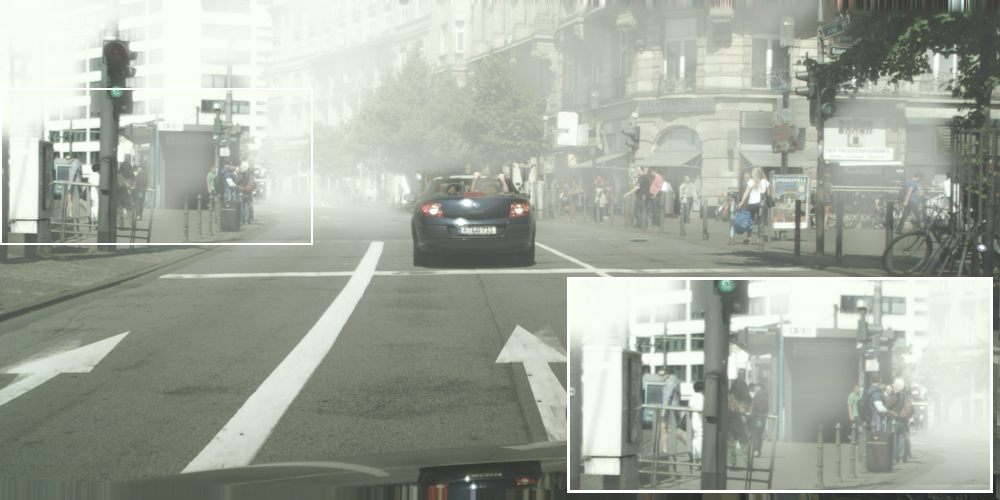} &
        \includegraphics[width=0.325\linewidth]{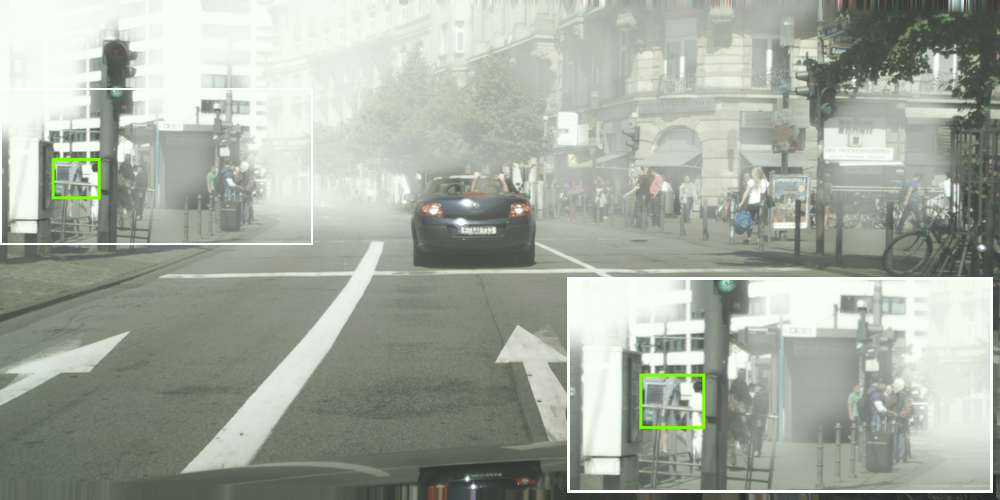} &
        \includegraphics[width=0.325\linewidth]{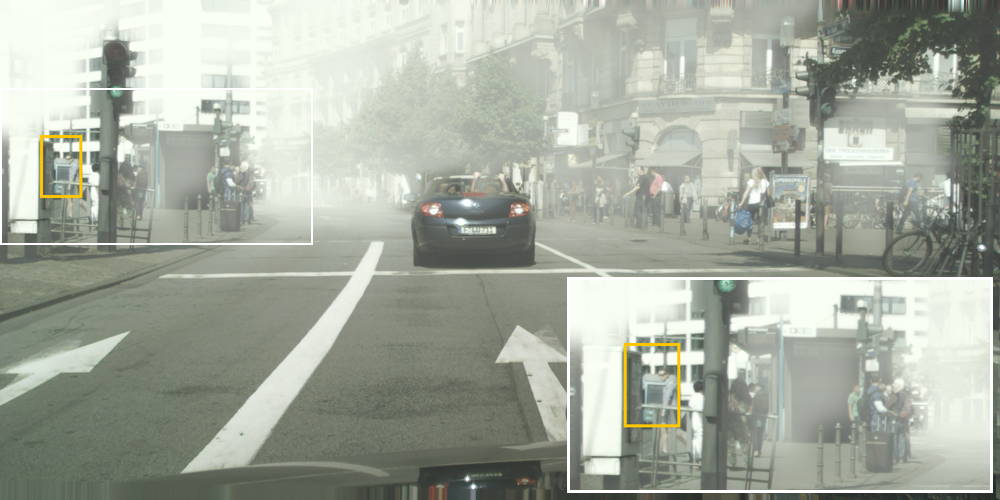} \\
    \end{tabular}
    \caption{Top to bottom: bus, rider, train. Left to right: source (baseline), \methname, GT. Insets magnify the region outlined in white.}
    \label{tab:qualitatives}
\end{table*}

\section{ABRA Algorithm}
\begin{algorithm*}
\caption{\methname: Aligned Basis Relocation for Adaptation}
\label{algo:transport}
\begin{algorithmic}[1]

\State \textbf{Input:} backbone weights $\theta_0$; source dataset $\mathcal{D}_S$; target dataset $\mathcal{D}_T$; class index $c$.

\smallskip
\State $\theta_S \gets \textsc{FineTune}(\theta_0, \tilde{\mathcal{D}}_S)$ \Comment{Source domain expert via objectification}
\State $U_S, \Sigma_S, V_S^\top \gets \textsc{SVD}(\theta_S)$ \Comment{Thin SVD at full rank, per adapted layer}

\smallskip
\State $\Delta\Sigma_c \gets \textsc{SVFT}(\theta_S,\mathcal{D}_S^{(c)}=\{(x_i,y_i=c)\})$
\Comment{Train class expert on source}

\smallskip
\State $\theta_T \gets \textsc{FineTune}(\theta_0, \tilde{\mathcal{D}}_T)$ \Comment{Target domain expert}
\State $U_T, \Sigma_T, V_T^\top \gets \textsc{SVD}(\theta_T)$

\smallskip
\State $L \gets U_T^\top U_S, \quad R \gets V_T^\top V_S$
\State $C_c \gets L\,\Delta\Sigma_c\,R^\top$ \Comment{Source residual in the target singular basis}
\State $\Sigma_T' \gets \Sigma_T + C_c$
\State \textbf{return} $\theta_T^{(c)} \gets U_T \Sigma_T' V_T^\top$ \Comment{Class-adapted target expert, $=\theta_T + P_U\tau_c P_V$}

\end{algorithmic}
\end{algorithm*}

\cref{algo:transport} outlines the procedure for \methname. First, the initial backbone weights $\theta_0$ are fine-tuned on the source dataset $\tilde{\mathcal{D}}_S$ and target dataset $\tilde{\mathcal{D}}_T$ to obtain the domain-specific experts $\theta_S$ and $\theta_T$, respectively. We then apply the thin Singular Value Decomposition (SVD) to both $\theta_S$ and $\theta_T$ to extract their singular vectors and values, $(U_S, \Sigma_S, V_S^\top)$ and $(U_T, \Sigma_T, V_T^\top)$.

To capture class-specific features, we perform Singular Vectors guided Fine-Tuning (SVFT) on the source expert $\theta_S$ using the source data subset restricted to class $c$, which isolates the class residual $\Delta\Sigma_c$. We then express this residual in the singular basis of the target expert: with $L = U_T^\top U_S$ and $R = V_T^\top V_S$, its coefficients are $C_c = L\,\Delta\Sigma_c\,R^\top$, which in weight space amounts to projecting the source residual onto the column and row spaces of $\theta_T$ (\cref{sup:optim}). Finally, these coefficients are added to the target singular values ($\Sigma_T' = \Sigma_T + C_c$), and the class-adapted target expert $\theta_T^{(c)}$ is reconstructed.

\section{Extended Performance Analysis}
\begin{table*}[t]
\centering
\caption{\textbf{Cityscapes $\rightarrow$ Foggy Cityscapes.} Comparison of AP$_{50}$ and AP$_{75}$ on unavailable categories, per-class protocol. \methname-$\alpha$ (oracle) uses, for all metrics, the $\alpha$ selected on mAP. \textbf{Bold}: best excluding the reference and the oracle.}
\label{tab:main_cityscapes}
\begingroup
\setlength{\tabcolsep}{3pt}
\resizebox{\linewidth}{!}{
\begin{tabular}{l *{10}{S} | *{2}{S}}
\toprule
& \multicolumn{2}{c}{\textbf{Bus}} & \multicolumn{2}{c}{\textbf{Motor}} & \multicolumn{2}{c}{\textbf{Rider}} & \multicolumn{2}{c}{\textbf{Train}} & \multicolumn{2}{c}{\textbf{Truck}} & \multicolumn{2}{c}{\textbf{Average}} \\
\cmidrule(lr){2-3}\cmidrule(lr){4-5}\cmidrule(lr){6-7}\cmidrule(lr){8-9}\cmidrule(lr){10-11}\cmidrule(lr){12-13}
\textbf{Method} & {AP$_{50}$} & {AP$_{75}$} & {AP$_{50}$} & {AP$_{75}$} & {AP$_{50}$} & {AP$_{75}$} & {AP$_{50}$} & {AP$_{75}$} & {AP$_{50}$} & {AP$_{75}$} & {AP$_{50}$} & {AP$_{75}$} \\
\midrule
\textit{Target fine-tuning (reference)} & {73.54} & {62.08} & {54.99} & {32.66} & {68.07} & {48.34} & {58.92} & {26.76} & {57.08} & {39.18} & {62.52} & {41.80} \\
\midrule
Zero shot & {61.03} & {53.47} & {41.83} & {21.54} & {31.72} & {19.77} & {44.02} & {8.03} & {41.88} & {32.70} & {44.10} & {27.10} \\
Source & {66.27} & {56.42} & {52.29} & {\textbf{29.77}} & {64.85} & {\textbf{47.24}} & {56.18} & {28.50} & {53.13} & {34.36} & {58.54} & {39.26} \\
FDA~\cite{yang2020fda} & {62.84} & {55.14} & {50.74} & {20.38} & {64.43} & {42.22} & {53.56} & {27.35} & {48.83} & {33.87} & {56.08} & {35.79} \\
Param$\Delta$~\cite{cao2025param} & {44.07} & {41.26} & {14.37} & {7.24} & {15.95} & {9.38} & {19.52} & {2.75} & {17.36} & {15.11} & {22.26} & {15.15} \\
\midrule
\rowcolor{customyellow!55}
\textbf{ABRA (ours)} & {\textbf{70.40}} & {\textbf{59.20}} & {\textbf{55.41}} & {26.01} & {\textbf{66.09}} & {45.58} & {\textbf{59.64}} & {\textbf{34.30}} & {\textbf{53.36}} & {\textbf{39.17}} & {\textbf{60.98}} & {\textbf{40.85}} \\
\midrule
\textit{ABRA-$\alpha$ (oracle)} & {70.59} & {58.94} & {55.17} & {26.33} & {66.09} & {45.58} & {54.53} & {49.64} & {54.31} & {39.77} & {60.14} & {44.05} \\
\bottomrule
\end{tabular}}
\endgroup
\end{table*}

\begin{table*}[t]
\centering
\caption{\textbf{SDGOD.} Comparison of AP$_{50}$ and AP$_{75}$ across challenging condition splits, per-class protocol. \methname-$\alpha$ (oracle) uses, for all metrics, the $\alpha$ selected on mAP. \textbf{Bold}: best excluding the reference and the oracle.}
\label{tab:sdgod}
\begingroup
\setlength{\tabcolsep}{5pt}
\resizebox{\linewidth}{!}{
\begin{tabular}{l *{8}{S} | *{2}{S}}
\toprule
& \multicolumn{2}{c}{\textbf{Day Foggy}} & \multicolumn{2}{c}{\textbf{Dusk Rainy}} & \multicolumn{2}{c}{\textbf{Night Clear}} & \multicolumn{2}{c}{\textbf{Night Rainy}} & \multicolumn{2}{c}{\textbf{Average}} \\
\cmidrule(lr){2-3}\cmidrule(lr){4-5}\cmidrule(lr){6-7}\cmidrule(lr){8-9}\cmidrule(lr){10-11}
\textbf{Method} & {AP$_{50}$} & {AP$_{75}$} & {AP$_{50}$} & {AP$_{75}$} & {AP$_{50}$} & {AP$_{75}$} & {AP$_{50}$} & {AP$_{75}$} & {AP$_{50}$} & {AP$_{75}$} \\
\midrule
\textit{Target fine-tuning (reference)} & {57.82} & {37.08} & {50.78} & {24.62} & {70.29} & {34.49} & {29.81} & {17.48} & {52.18} & {28.42} \\
\midrule
Zero shot & {41.10} & {26.91} & {34.93} & {20.30} & {49.63} & {25.44} & {13.60} & {10.23} & {34.82} & {20.72} \\
Source & {51.42} & {31.69} & {50.35} & {\textbf{26.49}} & {\textbf{66.64}} & {33.89} & {27.47} & {\textbf{14.61}} & {48.97} & {26.67} \\
FDA~\cite{yang2020fda} & {51.65} & {31.35} & {49.30} & {24.52} & {64.28} & {33.76} & {25.77} & {14.29} & {47.75} & {25.98} \\
Param$\Delta$~\cite{cao2025param} & {26.66} & {18.55} & {8.20} & {5.28} & {8.01} & {4.69} & {12.54} & {8.45} & {13.85} & {9.24} \\
\midrule
\rowcolor{customyellow!55}
\textbf{ABRA (ours)} & {\textbf{53.78}} & {\textbf{32.86}} & {\textbf{51.24}} & {26.41} & {66.12} & {\textbf{34.12}} & {\textbf{30.96}} & {14.28} & {\textbf{50.53}} & {\textbf{26.92}} \\
\midrule
\textit{ABRA-$\alpha$ (oracle)} & {54.35} & {34.62} & {49.67} & {26.70} & {65.65} & {34.37} & {31.38} & {16.01} & {50.26} & {27.92} \\
\bottomrule
\end{tabular}}
\endgroup
\end{table*}

In this supplementary section, we provide an extended analysis of the results presented in~\cref{tab:main_cityscapes} and~\cref{tab:sdgod}. \methname-$\alpha$ (oracle) uses, for all metrics, the $\alpha$ selected on mAP (\cref{tab:alpha_values}); as this $\alpha$ is not tuned on AP$_{50}$ or AP$_{75}$, the oracle can fall below plain \methname\ on these metrics.

Across both datasets, a critical indicator of localization precision is the stricter AP$_{75}$ metric. On Foggy Cityscapes, untuned \methname\ reaches an average AP$_{75}$ of $40.85$, above Source ($39.26$), FDA ($35.79$), and Param$\Delta$ ($15.15$), and $0.95$ below the Target fine-tuning reference ($41.80$). On SDGOD, it reaches the highest average AP$_{50}$ ($50.53$) and AP$_{75}$ ($26.92$) among the methods that respect our setting, while Param$\Delta$ degrades severely ($13.85$ and $9.24$). In the highly degraded \textit{Night Rainy} condition, \methname\ reaches an AP$_{50}$ of $30.96$, above every baseline and the reference.

\FloatBarrier
\section{Class Partitions and Dataset Statistics}
\label{sup:partitions}
\cref{tab:partitions} lists the class partitions and image counts of every benchmark. In all experiments, the objectified classes are the three most frequent classes of the source training set; they are shared by source and target domains and constitute the available target classes. Day-Clear is the source domain of every SDGOD split. \cref{tab:counts} reports per-class instance and image counts.
\begin{table*}[t]
\centering
\caption{\textbf{Class partitions and target splits.} Left: source classes $\mathcal{C}_S$, available (objectified) target classes $\mathcal{C}_T^{\text{avail}}$, and unavailable target classes $\mathcal{C}_T^{\text{unav}}$. Right: training and test images per target split; share of target training boxes from available classes; training images of $\theta_T$ (at least one available-class instance), and how many of them also contain an unavailable class. Sources: Cityscapes ($2{,}975$ training images) and SDGOD Day-Clear ($20{,}757$ training and $6{,}951$ test images).}
\label{tab:partitions}
\setlength{\tabcolsep}{3pt}
\resizebox{\linewidth}{!}{
\begin{tabular}{l p{3.6cm} p{1.7cm} p{2.1cm} || l r r r r r}
\toprule
\textbf{Benchmark} & $\mathcal{C}_S$ & $\mathcal{C}_T^{\text{avail}}$ & $\mathcal{C}_T^{\text{unav}}$ & \textbf{Target split} & Train img. & Test img. & Avail. boxes & $\theta_T$ img. & $\theta_T$ img. with unav. \\
\midrule
Cityscapes $\rightarrow$ Foggy Cityscapes & 8: person, rider, car, truck, bus, train, motorcycle, bicycle & car, person, bicycle & bus, motorcycle, rider, train, truck & Foggy Cityscapes & 2{,}975 & 500 & 93.2\% & 2{,}963 & 1{,}660 (56.0\%) \\
\midrule
\multirow{4}{*}{SDGOD: Day-Clear $\rightarrow$} & \multirow{4}{3.6cm}{7: bus, bike, car, motorcycle, person, rider, truck} & \multirow{4}{1.7cm}{car, person, truck} & \multirow{4}{2.1cm}{bus, bike, motorcycle, rider} & Day-Foggy & 2{,}812 & 963 & 86.4\% & 2{,}791 & 1{,}691 (60.6\%) \\
& & & & Dusk-Rainy & 2{,}611 & 890 & 96.7\% & 2{,}598 & 614 (23.6\%) \\
& & & & Night-Clear & 19{,}599 & 6{,}559 & 98.2\% & 19{,}372 & 2{,}204 (11.4\%) \\
& & & & Night-Rainy & 1{,}864 & 630 & 98.0\% & 1{,}850 & 239 (12.9\%) \\
\bottomrule
\end{tabular}}
\end{table*}

\begin{table*}[t]
\centering
\caption{\textbf{Per-class counts.} Instances (images containing at least one instance) per class and split. Available classes are listed first. Cityscapes and Foggy Cityscapes share the same images and annotations.}
\label{tab:counts}
\setlength{\tabcolsep}{3pt}
\resizebox{\linewidth}{!}{
\begin{tabular}{l cc || l cc cc cc cc cc}
\toprule
& \multicolumn{2}{c||}{\textbf{Cityscapes}} & & \multicolumn{2}{c}{\textbf{Day Clear}} & \multicolumn{2}{c}{\textbf{Day Foggy}} & \multicolumn{2}{c}{\textbf{Dusk Rainy}} & \multicolumn{2}{c}{\textbf{Night Clear}} & \multicolumn{2}{c}{\textbf{Night Rainy}} \\
\textbf{Class} & Train & Test & \textbf{Class} & Train & Test & Train & Test & Train & Test & Train & Test & Train & Test \\
\midrule
car & 27{,}155~(2{,}831) & 4{,}667~(478) & car & 237{,}746~(20{,}599) & 79{,}255~(6{,}897) & 24{,}868~(2{,}687) & 8{,}524~(917) & 25{,}432~(2{,}591) & 8{,}808~(882) & 180{,}701~(19{,}317) & 60{,}915~(6{,}465) & 16{,}062~(1{,}846) & 5{,}593~(624) \\
person & 17{,}994~(2{,}326) & 3{,}419~(402) & person & 29{,}978~(7{,}482) & 10{,}292~(2{,}513) & 15{,}954~(2{,}070) & 5{,}576~(707) & 3{,}760~(934) & 1{,}384~(321) & 13{,}104~(3{,}814) & 4{,}462~(1{,}286) & 1{,}100~(380) & 432~(130) \\
bicycle & 3{,}729~(1{,}538) & 1{,}175~(331) & truck & 12{,}650~(7{,}606) & 4{,}250~(2{,}550) & 538~(385) & 198~(138) & 1{,}655~(955) & 580~(326) & 3{,}646~(2{,}753) & 1{,}207~(927) & 373~(287) & 126~(97) \\
\midrule
rider & 1{,}807~(1{,}024) & 556~(254) & bus & 4{,}583~(3{,}438) & 1{,}530~(1{,}163) & 418~(288) & 136~(102) & 610~(461) & 210~(160) & 1{,}499~(1{,}259) & 513~(432) & 183~(163) & 65~(57) \\
motorcycle & 739~(511) & 149~(93) & bike & 2{,}549~(1{,}522) & 894~(521) & 3{,}672~(1{,}398) & 1{,}248~(481) & 249~(143) & 73~(54) & 1{,}052~(653) & 358~(228) & 91~(55) & 30~(21) \\
truck & 489~(363) & 93~(80) & motorcycle & 1{,}281~(923) & 468~(324) & 638~(457) & 273~(163) & 79~(65) & 31~(27) & 496~(410) & 169~(148) & 36~(30) & 13~(13) \\
bus & 385~(276) & 98~(75) & rider & 1{,}855~(1{,}431) & 663~(491) & 1{,}796~(952) & 567~(326) & 119~(103) & 50~(41) & 623~(529) & 218~(189) & 51~(45) & 20~(19) \\
train & 171~(145) & 23~(22) & & & & & & & & & & &  \\
\bottomrule
\end{tabular}}
\end{table*}

\section{Strict Domain Experts}
\label{sup:strict}
To quantify the effect of unlabeled instances of unavailable classes in the training images of $\theta_T$, we retrain strict domain experts after excluding all target images that contain any unavailable class, or only those containing the evaluated class (\cref{tab:strict}). Excluding these images changes \methname\ by only $0.70$ mAP on average, a difference that also reflects training on fewer images, while $\theta_T$ alone improves on the unavailable classes from $6.78$ to $15.84$ mAP, confirming that the unlabeled instances previously acted as background supervision against those classes.
\begin{table}[t]
\centering
\caption{\textbf{Strict domain experts.} \methname\ with target domain experts trained on all images, on the images without any unavailable class ($1{,}303$, $44.0\%$), or on the images without the evaluated class ($2{,}688$ for Bus, $2{,}819$ for Train). Cityscapes $\rightarrow$ Foggy Cityscapes, mAP, per-class protocol. $^*$Retrained; it differs from the $\theta_T$ of \cref{tab:transfer_origin} only by training variance.}
\label{tab:strict}
\setlength{\tabcolsep}{2.5pt}
\resizebox{\linewidth}{!}{
\begin{tabular}{l *{5}{S} | S}
\toprule
\textbf{$\theta_T$ trained on} & {\textbf{Bus}} & {\textbf{Motor}} & {\textbf{Rider}} & {\textbf{Train}} & {\textbf{Truck}} & {\textbf{Avg}} \\
\midrule
All images$^*$ & {55.70} & {29.83} & {42.32} & {35.14} & {37.41} & {40.08} \\
No unavailable class & {53.13} & {28.89} & {42.25} & {35.45} & {37.19} & {39.38} \\
No evaluated class & {53.97} & {--} & {--} & {34.69} & {--} & {--} \\
\bottomrule
\end{tabular}}
\end{table}

\section{Standard Evaluation Protocol}
\label{sup:standard}
\cref{tab:standard_protocol} reports per-class results on all $500$ test images of Foggy Cityscapes. Absolute scores decrease compared with the per-class protocol, most notably on Train, which appears in only $22$ of the $500$ images, but the ranking of the methods is unchanged: Target fine-tuning, \methname, Source, FDA, Zero shot, and Param$\Delta$. Only the oracle falls below plain \methname, as its $\alpha$ is selected under the per-class protocol.
\begin{table}[t]
\centering
\caption{\textbf{Standard protocol.} mAP on the unavailable classes of Cityscapes $\rightarrow$ Foggy Cityscapes, evaluated on all $500$ test images. \methname-$\alpha$ uses the $\alpha$ selected under the per-class protocol.}
\label{tab:standard_protocol}
\setlength{\tabcolsep}{2.5pt}
\resizebox{\linewidth}{!}{
\begin{tabular}{l *{5}{S} | S}
\toprule
\textbf{Method} & {\textbf{Bus}} & {\textbf{Motor}} & {\textbf{Rider}} & {\textbf{Train}} & {\textbf{Truck}} & {\textbf{Avg}} \\
\midrule
\textit{Target fine-tuning} & {44.41} & {22.58} & {40.89} & {17.53} & {30.75} & {31.23} \\
\midrule
Zero shot & {34.55} & {13.12} & {11.09} & {5.62} & {20.49} & {16.97} \\
Source & {42.36} & {20.54} & {\textbf{39.37}} & {11.89} & {28.51} & {28.53} \\
FDA~\cite{yang2020fda} & {40.09} & {15.67} & {35.86} & {7.56} & {24.19} & {24.67} \\
Param$\Delta$~\cite{cao2025param} & {23.94} & {1.54} & {5.57} & {1.92} & {5.43} & {7.68} \\
\midrule
\rowcolor{customyellow!55}
\textbf{ABRA (ours)} & {\textbf{46.22}} & {\textbf{21.47}} & {39.10} & {\textbf{13.53}} & {\textbf{29.91}} & {\textbf{30.05}} \\
\midrule
\textit{ABRA-$\alpha$ (oracle)} & {45.93} & {21.00} & {39.10} & {11.73} & {30.38} & {29.63} \\
\bottomrule
\end{tabular}}
\end{table}

\section{Masked Transports}
\label{sup:masks}
We attenuate or remove parts of the residual in the target basis: with $C = U_T^\top \tau_S V_T$, the masked transport is $\hat{\tau}_T = U_T (W \odot C) V_T^\top$, where $W = \mathbf{1}$ recovers \methname. Masks and thresholds are applied per layer. Band masks keep $|i - j| \le w$ and scale the entries outside by the given factor (zero by default); Gaussian and exponential masks weight entries by $\exp(-(i-j)^2/2s^2)$ and $\exp(-|i-j|/s)$; the leading-$128$ mask keeps the top-left $128 \times 128$ block, and has no effect on the $128 \times 256$ layers; the top-$5\%$ mask keeps the largest coefficients by magnitude; SCORE~\cite{chaves2026bridging} trimming keeps the diagonal and the off-diagonal entries within $1.96$ standard deviations of their mean. Direct addition adds $\tau_S$ to $\theta_T$ without any transport.

\cref{tab:masks_cs,tab:masks_df} report all masks per class. Removing most off-diagonal coefficients consistently degrades performance, whereas keeping the $5\%$ largest preserves it: class-specific information is concentrated in a small subset of large off-diagonal coefficients. On Day-Foggy, where \methname\ is $4.0$ mAP below the reference and the oracle $\alpha$ is below $1$ for every class, mild masks yield gains of up to $0.9$ mAP. However, the best mask on the available classes, the only selection our setting allows, does not transfer across benchmarks ($+0.70$ on Day-Foggy, $-0.45$ on Cityscapes), so we keep \methname\ unchanged.
\begin{table*}[t]
\centering
\caption{\textbf{Masked transports, Cityscapes $\rightarrow$ Foggy Cityscapes.} mAP per class, per-class protocol. Rows below the rule are sorted by the average on the unavailable classes. Rows ran on different GPU models; on the same checkpoint, differences reach $0.14$ mAP on a class average, so differences below about $0.2$ mAP are not effects.}
\label{tab:masks_cs}
\setlength{\tabcolsep}{4pt}
\resizebox{\linewidth}{!}{
\begin{tabular}{l *{5}{S} | S || *{3}{S} | S}
\toprule
& \multicolumn{6}{c||}{\textbf{Unavailable classes}} & \multicolumn{4}{c}{\textbf{Available classes}} \\
\textbf{Mask $W$} & {\textbf{Bus}} & {\textbf{Motorcycle}} & {\textbf{Rider}} & {\textbf{Train}} & {\textbf{Truck}} & {\textbf{Avg}} & {\textbf{Car}} & {\textbf{Person}} & {\textbf{Bicycle}} & {\textbf{Avg}} \\
\midrule
None (\methname) & {57.22} & {30.01} & {42.29} & {34.97} & {38.12} & {40.52} & {50.83} & {37.40} & {31.94} & {40.06} \\
Direct addition & {57.27} & {30.09} & {42.36} & {35.29} & {38.32} & {40.66} & {50.85} & {37.43} & {31.93} & {40.07} \\
\midrule
$5\%$ largest coefficients & {57.17} & {29.98} & {42.26} & {34.22} & {38.01} & {40.33} & {50.79} & {37.48} & {32.08} & {40.12} \\
Band $|i-j|\le4$ & {56.87} & {30.02} & {42.06} & {33.09} & {38.32} & {40.07} & {50.90} & {37.54} & {32.12} & {40.19} \\
Band $|i-j|\le2$, outside $\times0.5$ & {56.80} & {30.15} & {42.18} & {32.90} & {38.20} & {40.04} & {50.81} & {37.37} & {31.93} & {40.04} \\
SVFT band $|i-j|\le2$ & {56.51} & {28.91} & {41.56} & {32.24} & {37.88} & {39.42} & {50.71} & {37.29} & {31.86} & {39.95} \\
Gaussian, $s=2$ & {55.27} & {26.87} & {39.45} & {30.36} & {36.70} & {37.73} & {50.76} & {36.97} & {31.95} & {39.89} \\
Leading $128$ directions & {54.35} & {25.02} & {37.35} & {30.48} & {35.92} & {36.62} & {50.29} & {36.51} & {31.13} & {39.31} \\
Diagonal, outside $\times0.5$ & {53.15} & {24.95} & {34.54} & {28.73} & {35.11} & {35.30} & {50.62} & {36.63} & {31.92} & {39.72} \\
Band $|i-j|\le1$ & {52.38} & {21.54} & {30.04} & {26.94} & {34.37} & {33.05} & {50.29} & {36.19} & {30.75} & {39.08} \\
Gaussian, $s=1$ & {49.24} & {18.62} & {21.98} & {25.58} & {32.29} & {29.54} & {50.09} & {35.91} & {30.66} & {38.89} \\
Diagonal, outside $\times0.25$ & {47.84} & {17.24} & {17.94} & {24.35} & {30.22} & {27.52} & {49.93} & {35.75} & {30.69} & {38.79} \\
Exponential, $s=1$ & {46.87} & {16.10} & {16.50} & {22.52} & {29.11} & {26.22} & {49.77} & {35.55} & {29.98} & {38.43} \\
SCORE trimming ($\mu\pm1.96\sigma$) & {42.80} & {12.56} & {11.59} & {14.87} & {22.71} & {20.91} & {48.61} & {34.81} & {28.30} & {37.24} \\
Diagonal only & {38.93} & {9.35} & {10.30} & {10.09} & {18.12} & {17.36} & {48.32} & {34.43} & {26.36} & {36.37} \\
\bottomrule
\end{tabular}}
\end{table*}
\begin{table*}[t]
\centering
\caption{\textbf{Masked transports, SDGOD Day-Foggy.} mAP per class, per-class protocol. Rows below the rule are sorted by the average on the unavailable classes. Rows ran on different GPU models; on the same checkpoint, differences reach $0.14$ mAP on a class average, so differences below about $0.2$ mAP are not effects.}
\label{tab:masks_df}
\setlength{\tabcolsep}{4pt}
\resizebox{\linewidth}{!}{
\begin{tabular}{l *{4}{S} | S || *{3}{S} | S}
\toprule
& \multicolumn{5}{c||}{\textbf{Unavailable classes}} & \multicolumn{4}{c}{\textbf{Available classes}} \\
\textbf{Mask $W$} & {\textbf{Bus}} & {\textbf{Bike}} & {\textbf{Motorcycle}} & {\textbf{Rider}} & {\textbf{Avg}} & {\textbf{Car}} & {\textbf{Person}} & {\textbf{Truck}} & {\textbf{Avg}} \\
\midrule
None (\methname) & {46.75} & {23.37} & {25.74} & {33.51} & {32.34} & {47.34} & {34.58} & {34.76} & {38.89} \\
Direct addition & {46.76} & {23.33} & {25.63} & {33.62} & {32.34} & {47.22} & {34.59} & {34.74} & {38.85} \\
\midrule
Band $|i-j|\le2$, outside $\times0.5$ & {47.08} & {23.99} & {27.28} & {34.59} & {33.24} & {47.74} & {34.82} & {34.90} & {39.16} \\
Band $|i-j|\le4$ & {46.72} & {23.65} & {28.00} & {34.43} & {33.20} & {47.87} & {34.74} & {33.32} & {38.64} \\
SVFT band $|i-j|\le2$ & {46.44} & {24.17} & {27.81} & {34.07} & {33.12} & {47.90} & {34.67} & {33.24} & {38.61} \\
Leading $128$ directions & {47.10} & {23.63} & {27.64} & {33.78} & {33.04} & {48.24} & {34.72} & {35.72} & {39.56} \\
$5\%$ largest coefficients & {47.04} & {23.55} & {27.03} & {34.15} & {32.94} & {47.68} & {34.78} & {34.09} & {38.85} \\
Gaussian, $s=2$ & {46.04} & {24.04} & {28.43} & {32.75} & {32.81} & {48.18} & {34.65} & {32.70} & {38.51} \\
Diagonal, outside $\times0.5$ & {46.04} & {24.08} & {28.16} & {32.49} & {32.69} & {48.47} & {34.73} & {33.63} & {38.94} \\
Band $|i-j|\le1$ & {44.36} & {23.06} & {27.06} & {26.02} & {30.13} & {48.46} & {34.20} & {30.11} & {37.59} \\
Diagonal, outside $\times0.25$ & {43.84} & {22.73} & {26.50} & {21.25} & {28.58} & {48.55} & {34.27} & {28.04} & {36.95} \\
Gaussian, $s=1$ & {43.23} & {22.40} & {25.99} & {20.60} & {28.05} & {48.54} & {34.18} & {27.21} & {36.64} \\
SCORE trimming ($\mu\pm1.96\sigma$) & {43.48} & {21.59} & {26.36} & {12.47} & {25.98} & {48.64} & {33.93} & {26.08} & {36.22} \\
Exponential, $s=1$ & {42.03} & {21.33} & {24.85} & {15.25} & {25.87} & {48.55} & {34.01} & {24.80} & {35.78} \\
Diagonal only & {37.81} & {14.67} & {18.68} & {7.82} & {19.75} & {48.45} & {33.58} & {18.93} & {33.65} \\
\bottomrule
\end{tabular}}
\end{table*}

\section{Number of Objectified Classes}
\label{sup:k}
The top three classes cover $93.2\%$ of the source training boxes on Cityscapes and $96.5\%$ on SDGOD Day-Clear. One or two classes cover $51.8\%$ and $86.0\%$ on Cityscapes, while a fourth class adds only $3.4\%$ ($1.6\%$ on SDGOD). \cref{tab:k_ablation} reports \methname\ on Bus and Train with domain experts retrained for $k \in \{1, 2, 3\}$: no $k$ is best on both classes, and their average varies by less than $2$ mAP, so we retain $k = 3$ for its box coverage.
\begin{table}[t]
\centering
\caption{\textbf{Number of objectified classes $k$.} \methname\ on Bus and Train, Cityscapes $\rightarrow$ Foggy Cityscapes, mAP, per-class protocol. Domain experts are retrained for this ablation, so the $k=3$ column differs from \cref{tab:main_results_merged} by training variance.}
\label{tab:k_ablation}
\setlength{\tabcolsep}{4pt}
\begin{tabular}{l c c c}
\toprule
& $k=1$ & $k=2$ & $k=3$ \\
& (car) & (+ person) & (+ bicycle) \\
\midrule
Bus & 55.25 & 53.32 & 56.41 \\
Train & 31.22 & 36.96 & 32.66 \\
\midrule
Average & 43.24 & 45.14 & 44.54 \\
\bottomrule
\end{tabular}
\end{table}

\clearpage

\bibliographystyle{cas-model2-names}

\bibliography{cas-refs}

@string{cvpr = {Proceedings of the IEEE conference on Computer Vision and Pattern Recognition}}

@string{eccv = {Proceedings of the European Conference on Computer Vision}}

@string{iccv = {IEEE International Conference on Computer Vision}}

@string{nips = {Neural Information Processing Systems}}

@string{neuripsw = {Neural Information Processing Systems Workshops}}

@string{iclr = {International Conference on Learning Representations}}

@string{iclr = {International Conference on Learning Representations Workshop}}

@string{icml = {International Conference on Machine Learning}}

@string{tmlr = {Transactions on Machine Learning Research}}

@string{ijcv = {International Journal of Computer Vision}}

@string{tpami = {IEEE Transactions on Pattern Analysis and Machine Intelligence}}

@string{tcsvt = {IEEE Transactions on Circuits and Systems for Video Technology}}

@string{nature = {Nature}}

@string{cvpr = {CVPR}}

@string{eccv = {ECCV}}

@string{iccv = {ICCV}}

@string{nips = {NeurIPS}}

@string{neuripsw = {NeurIPS Workshops}}

@string{iclr = {ICLR}}

@string{icml = {ICML}}

@string{tmlr = {Trans. Mach. Learn. Res.}}

@string{ijcv = {Int. J. Comput. Vis.}}

@string{tpami = {IEEE Trans. Pattern Anal. Mach. Intell.}}

@string{tcsvt = {IEEE Trans. Circuits Syst. Video Technol.}}

@inproceedings{liu2024grounding,
  title={Grounding {DINO}: Marrying dino with grounded pre-training for open-set object detection},
  author={Liu, Shilong and Zeng, Zhaoyang and Ren, Tianhe and Li, Feng and Zhang, Hao and Yang, Jie and Jiang, Qing and Li, Chunyuan and Yang, Jianwei and Su, Hang and others},
  booktitle=eccv,
  year={2024}
}

@article{lingam2024svft,
  title={Svft: Parameter-efficient fine-tuning with singular vectors},
  author={Lingam, Vijay Chandra and Neerkaje, Atula and Vavre, Aditya and Shetty, Aneesh and Gudur, Gautham Krishna and Ghosh, Joydeep and Choi, Eunsol and Dimakis, Alex and Bojchevski, Aleksandar and Sanghavi, Sujay},
  journal=nips,
  year={2024}
}

@article{rinaldi2025update,
  title={Update Your Transformer to the Latest Release: Re-Basin of Task Vectors},
  author={Rinaldi, Filippo and Capitani, Giacomo and Bonicelli, Lorenzo and Crisostomi, Donato and Bolelli, Federico and Ficarra, Elisa and Rodola, Emanuele and Calderara, Simone and Porrello, Angelo},
  journal=icml,
  year={2025}
}

@article{hu2022lora,
  title={Lora: Low-rank adaptation of large language models.},
  author={Hu, Edward J and Shen, Yelong and Wallis, Phillip and Allen-Zhu, Zeyuan and Li, Yuanzhi and Wang, Shean and Wang, Lu and Chen, Weizhu and others},
  journal=iclr,
  year={2022}
}

@inproceedings{zhang2023adalora,
  title={Adalora: Adaptive budget allocation for parameter-efficient fine-tuning},
  author={Zhang, Qingru and Chen, Minshuo and Bukharin, Alexander and Karampatziakis, Nikos and He, Pengcheng and Cheng, Yu and Chen, Weizhu and Zhao, Tuo},
  booktitle=iclr,
  year={2023}
}

@inproceedings{gu2022open,
  title={Open-vocabulary Object Detection via Vision and Language Knowledge Distillation},
  author={Gu, Xiuye and Lin, Tsung-Yi and Kuo, Weicheng and Cui, Yin},
  booktitle=iclr,
  year={2022}
}

@article{cappellino2025dithub,
   title={DitHub: A Modular Framework for Incremental Open-Vocabulary Object Detection},
   author={Cappellino, Chiara and Mancusi, Gianluca and Mosconi, Matteo and Porrello, Angelo and Calderara, Simone and Cucchiara, Rita},
   journal=nips,
   year={2025}
}

@inproceedings{zhong2022regionclip,
  title={Regionclip: Region-based language-image pretraining},
  author={Zhong, Yiwu and Yang, Jianwei and Zhang, Pengchuan and Li, Chunyuan and Codella, Noel and Li, Liunian Harold and Zhou, Luowei and Dai, Xiyang and Yuan, Lu and Li, Yin and others},
  booktitle=cvpr,
  year={2022}
}

@inproceedings{zhou2022detecting,
  title={Detecting twenty-thousand classes using image-level supervision},
  author={Zhou, Xingyi and Girdhar, Rohit and Joulin, Armand and Kr{\"a}henb{\"u}hl, Philipp and Misra, Ishan},
  booktitle=eccv,
  year={2022},
}

@inproceedings{li2021grounded,
  title={Grounded language-image pre-training},
  author={Li, Liunian Harold and Zhang, Pengchuan and Zhang, Haotian and Yang, Jianwei and Li, Chunyuan and Zhong, Yiwu and Wang, Lijuan and Yuan, Lu and Zhang, Lei and Hwang, Jenq-Neng and others},
  booktitle=cvpr,
  year={2022}
}

@inproceedings{minderer2022simple,
  title={Simple open-vocabulary object detection},
  author={Minderer, Matthias and Gritsenko, Alexey and Stone, Austin and Neumann, Maxim and Weissenborn, Dirk and Dosovitskiy, Alexey and Mahendran, Aravindh and Arnab, Anurag and Dehghani, Mostafa and Shen, Zhuoran and others},
  booktitle=eccv,
  year={2022},
}

@article{pfeiffer2023modular,
  title={Modular Deep Learning},
  author={Pfeiffer, Jonas and Ruder, Sebastian and Vuli{\'c}, Ivan and Ponti, Edoardo},
  journal=tmlr,
  year=2023
}

@article{li2024vb,
  title={Vb-lora: Extreme parameter efficient fine-tuning with vector banks},
  author={Li, Yang and Han, Shaobo and Ji, Shihao},
  journal=nips,
  year={2024}
}

@inproceedings{kopiczko2024vera,
  title={Vera: Vector-based random matrix adaptation},
  author={Kopiczko, Dawid J and Blankevoort, Tijmen and Asano, Yuki M},
  booktitle=iclr,
  year={2024}
}

@inproceedings{liu2024dora,
  title={Dora: Weight-decomposed low-rank adaptation},
  author={Liu, Shih-Yang and Wang, Chien-Yi and Yin, Hongxu and Molchanov, Pavlo and Wang, Yu-Chiang Frank and Cheng, Kwang-Ting and Chen, Min-Hung},
  booktitle=icml,
  year={2024}
}

@article{ainsworth2022git,
  title={Git re-basin: Merging models modulo permutation symmetries},
  author={Ainsworth, Samuel K and Hayase, Jonathan and Srinivasa, Siddhartha},
  journal=iclr,
  year={2023}
}

@inproceedings{zang2022open,
  title={Open-vocabulary detr with conditional matching},
  author={Zang, Yuhang and Li, Wei and Zhou, Kaiyang and Huang, Chen and Loy, Chen Change},
  booktitle=eccv,
  year={2022},
}

@inproceedings{radford2021learning,
  title={Learning transferable visual models from natural language supervision},
  author={Radford, Alec and Kim, Jong Wook and Hallacy, Chris and Ramesh, Aditya and Goh, Gabriel and Agarwal, Sandhini and Sastry, Girish and Askell, Amanda and Mishkin, Pamela and Clark, Jack and others},
  booktitle=icml,
  year={2021}
}

@inproceedings{deng2024zero,
 title={Zero-shot Generalizable Incremental Learning for Vision-Language Object Detection},
 author={Deng, Jieren and Zhang, Haojian and Ding, Kun and Hu, Jianhua and Zhang, Xingxuan and Wang, Yunkuan},
 booktitle=nips,
 year={2024}
}

@article{van2022three,
  title={Three types of incremental learning},
  author={Van de Ven, Gido M and Tuytelaars, Tinne and Tolias, Andreas S},
  journal={Nature Machine Intelligence},
  year={2022}
}

@article{bansal2021revisiting,
  title={Revisiting model stitching to compare neural representations},
  author={Bansal, Yamini and Nakkiran, Preetum and Barak, Boaz},
  journal=nips,
  year={2021}
}

@inproceedings{hernandez2022model,
  title={Model Stitching: Looking For Functional Similarity Between Representations},
  author={Hernandez, Adriano and Dangovski, Rumen and Lu, Peter Y and Soljacic, Marin},
  booktitle=neuripsw,
  year={2022}
}

@inproceedings{pan2023stitchable,
  title={Stitchable neural networks},
  author={Pan, Zizheng and Cai, Jianfei and Zhuang, Bohan},
  booktitle=cvpr,
  year={2023}
}

@inproceedings{stoica2024zipit,
  title={ZipIt! Merging Models from Different Tasks without Training},
  author={Stoica, George and Bolya, Daniel and Bjorner, Jakob Brandt and Ramesh, Pratik and Hearn, Taylor and Hoffman, Judy},
  booktitle=iclr,
  year={2024}
}

@article{meng2024pissa,
  title={Pissa: Principal singular values and singular vectors adaptation of large language models},
  author={Meng, Fanxu and Wang, Zhaohui and Zhang, Muhan},
  journal=nips,
  year={2024}
}

@inproceedings{ilharco2023editing,
  title={Editing models with task arithmetic},
  author={Ilharco, Gabriel and Ribeiro, Marco Tulio and Wortsman, Mitchell and Schmidt, Ludwig and Hajishirzi, Hannaneh and Farhadi, Ali},
  booktitle=iclr,
  year={2023}
}

@article{chaves2026bridging,
  title={Bridging Domains through Subspace-Aware Model Merging},
  author={Chaves, Levy and Zhou, Chao and Burkholz, Rebekka and Valle, Eduardo and Avila, Sandra},
  journal=cvpr,
  year={2026}
}

@article{ortiz2023task,
  title={Task arithmetic in the tangent space: Improved editing of pre-trained models},
  author={Ortiz-Jimenez, Guillermo and Favero, Alessandro and Frossard, Pascal},
  journal=nips,
  year={2023}
}

@article{cao2025param,
  title={Param Delta for Direct Weight Mixing: Post-Train Large Language Model at Zero Cost},
  author={Cao, Sheng and Wu, Mingrui and Prasad, Karthik and Tian, Yuandong and Liu, Zechun},
  journal=iclr,
  year={2025}
}

@ARTICLE{10466767,
  author={Deng, Jinhong and Li, Wen and Duan, Lixin},
  journal=tcsvt, 
  title={Balanced Teacher for Source-Free Object Detection}, 
  year={2024},
}

@article{Sangineto_2019,
   title={Self Paced Deep Learning for Weakly Supervised Object Detection},
   journal=tpami,
   author={Sangineto, Enver and Nabi, Moin and Culibrk, Dubravko and Sebe, Nicu},
   year={2019},
}

@inproceedings{Cordts2016Cityscapes,
    title={The Cityscapes Dataset for Semantic Urban Scene Understanding},
    author={Cordts, Marius and Omran, Mohamed and Ramos, Sebastian and Rehfeld, Timo and Enzweiler, Markus and Benenson, Rodrigo and Franke, Uwe and Roth, Stefan and Schiele, Bernt},
    booktitle=cvpr,
    year={2016}
}

@article{SDV18,
  author = {Sakaridis, Christos and Dai, Dengxin and Van Gool, Luc},
  title = {Semantic Foggy Scene Understanding with Synthetic Data},
  journal = ijcv,
  year = {2018},
}

@inproceedings{wu2022single,
  title={Single-Domain Generalized Object Detection in Urban Scene via Cyclic-Disentangled Self-Distillation},
  author={Wu, Aming and Deng, Cheng},
  booktitle=cvpr,
  year={2022}
}

@inproceedings{10.5555/3294771.3294885,
    author = {Tarvainen, Antti and Valpola, Harri},
    title = {Mean teachers are better role models: Weight-averaged consistency targets improve semi-supervised deep learning results},
    year = {2017},
    booktitle=nips
}

@article{tang2025sourcefreedomainadaptiveobject,
      title={Source-Free Domain Adaptive Object Detection with Semantics Compensation}, 
      author={Song Tang and Jiuzheng Yang and Mao Ye and Boyu Wang and Yan Gan and Xiatian Zhu},
      year={2025},
      journal={arXiv preprint arXiv:2410.05557},
}

@inproceedings{kennerley2024catexploitinginterclassdynamics,
      title={CAT: Exploiting Inter-Class Dynamics for Domain Adaptive Object Detection}, 
      author={Mikhail Kennerley and Jian-Gang Wang and Bharadwaj Veeravalli and Robby T. Tan},
      year={2024},
      booktitle=cvpr,
}

@inproceedings{zhang2018mixupempiricalriskminimization,
      title={mixup: Beyond Empirical Risk Minimization}, 
      author={Hongyi Zhang and Moustapha Cisse and Yann N. Dauphin and David Lopez-Paz},
      year={2018},
      booktitle=iclr 
}

@inproceedings{mattolin2022confmixunsuperviseddomainadaptation,
      title={ConfMix: Unsupervised Domain Adaptation for Object Detection via Confidence-based Mixing}, 
      author={Giulio Mattolin and Luca Zanella and Elisa Ricci and Yiming Wang},
      year={2023},
      booktitle={Winter Conference on Applications of Computer Vision},
}

@inproceedings{yang2020fda,
  title={Fda: Fourier domain adaptation for semantic segmentation},
  author={Yang, Yanchao and Soatto, Stefano},
  booktitle=cvpr,
  year={2020}
}

@InProceedings{Yuan_2025_ICCV,
    author    = {Yuan, Yuxuan and Tang, Luyao and Chen, Yixin and Chen, Chaoqi and Huang, Yue and Ding, Xinghao},
    title     = {ASGS: Single-Domain Generalizable Open-Set Object Detection via Adaptive Subgraph Searching},
    booktitle = iccv,
    year      = {2025},
}

\end{document}